\documentclass[10pt,journal,compsoc]{IEEEtran}
\usepackage[numbers]{natbib}
\usepackage{url}

\usepackage{mathtools}
\usepackage{graphicx}
\usepackage{algorithm, algpseudocode}
\usepackage{subcaption}
\usepackage{booktabs}
\usepackage{amsfonts}
\usepackage{multirow}
\usepackage{enumitem}

\begin{document}
%
\title{A Differentiable Framework for End-to-End Learning of Hybrid Structured Compression}
%
%

\author{Moonjung~Eo, Suhyun~Kang, 
and~Wonjong~Rhee,~\IEEEmembership{Fellow,~IEEE}
\IEEEcompsocitemizethanks{\IEEEcompsocthanksitem The authors are with Deep Representation Learning Research Group, Department of Intelligence and Information, Seoul National University, Seoul, South Korea.\protect\\
E-mail: \{eod87,~su\_hyun,~wrhee\} @snu.ac.kr
\IEEEcompsocthanksitem Wonjong Rhee is also with IPAI (Interdisciplinary Program in Artificial Intelligence) and AI Institute at Seoul National University.}
\thanks{Manuscript received ...}}

%
%

\markboth{Journal of \LaTeX\ Class Files,~Vol.~14, No.~8, August~2015}%
{Shell \MakeLowercase{\textit{et al.}}: Bare Demo of IEEEtran.cls for Computer Society Journals}
%

\IEEEtitleabstractindextext{%
\begin{abstract}
Filter pruning and low-rank decomposition are two of the foundational techniques for structured compression. Although recent efforts have explored hybrid approaches aiming to integrate the advantages of both techniques, their performance gains have been modest at best. In this study, we develop a \textit{Differentiable Framework~(DF)} that can express filter selection, rank selection, and budget constraint into a single analytical formulation. Within the framework, we introduce DML-S for filter selection, integrating scheduling into existing mask learning techniques. 
Additionally, we present DTL-S for rank selection, utilizing a singular value thresholding operator.
The framework with DML-S and DTL-S offers a hybrid structured compression methodology that facilitates end-to-end learning through gradient-base optimization.  
Experimental results demonstrate the efficacy of DF, surpassing state-of-the-art structured compression methods. Our work establishes a robust and versatile avenue for advancing structured compression techniques.
\end{abstract}

}

\maketitle

\IEEEdisplaynontitleabstractindextext

%
\IEEEpeerreviewmaketitle

%
\IEEEpeerreviewmaketitle

\section{Introduction}
\label{sec:intro}

\IEEEPARstart{D}{eep neural networks (DNNs)} has achieved remarkable success across various domains but are often associated with high computational costs and memory demands. Structured compression (SC) techniques, such as filter pruning and low-rank decomposition, have emerged as promising solutions to tackle these challenges by reducing model complexity without compromising performance~\cite{idelbayev2020low, li2020group, ruan2020edp, shang2022neural,sui2021chip,zhuang2018discrimination}. Unlike unstructured compression techniques, SC can be easily applied because it does not require a special hardware~\cite{han2016eie} or software~\cite{park2016faster} accelerator.

\textit{Filter pruning} aims to eliminate redundant filters with negligible contributions to the network's performance~\cite{he2018soft, huang2018data, liu2019metapruning,luo2020autopruner}. 
On the other hand, \textit{low-rank decomposition} approximates weight matrices with lower-rank counterparts, leveraging the correlation among filter weights~\cite{idelbayev2020low,kim2019efficient, liebenwein2021compressing, yaguchi2019decomposable}. 
\textit{Hybrid structured compression} simultaneously pursues the goals of filter pruning and low-rank decomposition. Its objective is to identify optimal rank and filters within the weight filter that lead to desired compressed architecture of the network while preserving the network's performance.
\begin{figure}[t]
    \centering
    \begin{subfigure}{\linewidth}
        \centering
        \includegraphics[width=0.95\linewidth]{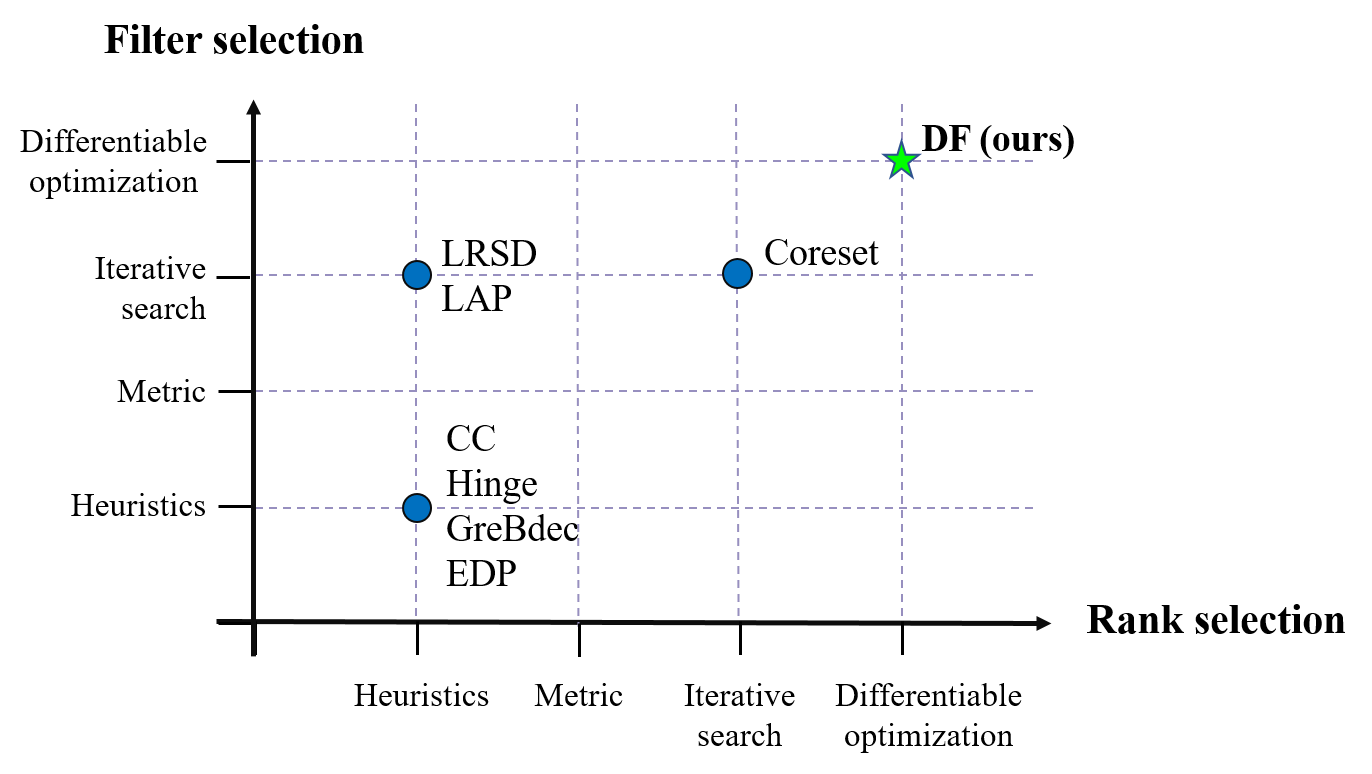}
    \end{subfigure}
    \hfill
\caption{The filter selection and rank selection methods of the latest hybrid compression algorithms are shown. Blue circles denote the previously developed hybrid compression methods and the green star denotes our method.} 
\label{fig:process_comparison}
\end{figure}

In hybrid structured compression, the key problem is the selection of the suitable filters and ranks for each layer of a network.
This problem falls under the category of Combinatorial Optimization Problem (COP)~\cite{cook1995combinatorial}. Moreover, the addition of a resource budget constraint transforms it into a more complex form known as a Constrained Combinatorial Optimization Problem (CCOP). Since CCOP is known as an NP-hard problem, previous studies on hybrid structured compression have relied on classical approaches, such as heuristics~\cite{yu2017compressing,li2020group,ruan2020edp,li2021towards}, iterative search techniques~\cite{dubey2018coreset}, or a combination of both~\cite{guo2019compressing,chen2020deep}, to address the CCOP problem as summarized in Figure~\ref{fig:process_comparison}. 
However, these approaches face difficulties in searching for the optimal filters and ranks in each layer because of the exponential growth of the solution space with the number of layers~\cite{idelbayev2020low}.
As the network complexity increases, exploring all possible combinations becomes computationally infeasible, making it challenging to find the solution efficiently. 
In a network with $K$-layers of weights, each with a maximum rank of $R_k$ and a number of filters of $F_k$, the total number of possible combinations of filters and ranks is given by $\prod^K_{k=1}2^{F_k}R_{k}$. 
In large networks, where $F_k$ and $R_k$ can be in the range of thousands, this leads to immensely large solution space.
While gradient-based optimization is another approach to effectively solve the problem, it has received much less attention. This is because the original discrete form of the CCOP does not allow a direct utilization of the gradient-based optimization methods.

In this study, we focus on applying gradient-based optimization to solve the CCOP to select the optimal filters and ranks under a budget constraint. 
We propose \textit{Differentiable Framework}~(DF), designed to facilitate the end-to-end learning of rank and filters for hybrid structured compression.
It is a generic framework that reparametrizes the original discrete solution space into a continuous latent space, where the problem can be approximately solved by gradient-based optimization in an end-to-end fashion. First, we mathematically formulate the CCOP to select the optimal filters and ranks under a budget constraint. Then, we reparameterize the discrete parameters to continuous parameters and approximate the budget constraint to penalty term, resulting in approximating the CCOP as a continuous optimization problem. Finally, we learn the parameters in an end-to-end manner based on gradient-based optimization. 
As illustrated in Figure~\ref{fig:process_comparison}),
DF primarily differs from the previous hybrid methods because it adopts an end-to-end learning approach. 
We conduct extensive experiments to demonstrate the effectiveness of DF compared to existing compression techniques. 

The remainder of this paper is organized as follows: Section 2 provides a detailed review of related works on model compression, filter pruning, and low-rank decomposition. Backgrounds of our method are introduced in Section 3. In Section 4, we present our gradient-based approach, outlining the formulation of the unified criterion and optimization procedure. Section 5 presents the experimental results, illustrating the efficacy and benefits of our proposed method. Finally, in Section 6, we conclude our findings and discuss potential directions for future research.

\section{Related works}
\label{sec:related_works}
\subsection{Hybrid compression methods}
Previous research~\cite{dubey2018coreset, chen2020deep} has proposed separate compression stages to integrate multiple compression techniques. These stages sequentially adopt one compression technique in each step and ignore the interrelations of the different compression methods. For instance, in \cite{dubey2018coreset}, filter pruning is conducted first to reduce the weights, and the weights are then decomposed using a coreset-based decomposition technique. In addition, several compression-aware training approaches have been proposed using regularization to make a network compression-friendly~\cite{chen2019exploiting, ruan2020edp, guo2019compressing, li2020group}. For example, in \cite{li2020group}, they first introduce a sparsity-inducing matrix at each weight and then impose group sparsity constraints during training. However, determining a balance between compression rate and accuracy is challenging under the desired compression rate with these compression-aware methods. Recently, \cite{li2022heuristic} proposes a collaborative compression method to employ the global compression rate optimization method to obtain the compression rate of each layer and adopt a multi-step heuristic removal strategy. Our hybrid compression method does not need heuristics on selecting filters and ranks, and does not require compression-aware regularized training.

\subsection{Rank selection for low-rank decomposition}

In the early investigations of low-rank decomposition, researchers employed heuristic strategies, such as a greedy approach or a thresholding method, to determine ranks in advance~\cite{kim2015compression, kim2019efficient, wen2017coordinating, xu2020trp}. Subsequently, they observed a correlation between the energy (the sum of singular values) and DNN performance. To preserve performance while minimizing the impact on energy, rank selection methods were proposed~\cite{zhang2015accelerating, alvarez2017compression, li2018constrained}. Some researchers formulated rank selection as optimization problems, incorporating the singular values in their formulations~[16, 1, 17, 18]. However, these methods have limitations as they rely on heuristics or approximations to solve the overall problem and are restricted to layer-wise rank selection for applying their optimization formulations, failing to achieve globally optimal configurations. An alternative line of research involves using a genetic algorithm to determine the proper rank configuration~[26]. Inspired by neural architecture search, this strategy employs heuristically determined parameters and entails an expensive overall iterative search process. 
Distinct from the previous research, we are the first to transform the rank selection problem into a fully differentiable optimization problem. 

\subsection{Filter selection for filter pruning}
Previous research on filter pruning can be categorized into two main groups based on how the importance of each filter is determined: criteria-based and learning-based methods. Criteria-based methods rely on specific metrics to evaluate the importance of filters within a layer. 
These metrics can include $L_2$-norm/$L_1$-norm of the filter weights~\cite{li2016pruning}, geometric median of the filters~\cite{he2019filter}, or error contribution towards the output of the current layer~\cite{luo2017thinet, he2017channel}. 
Filters with scores below a certain threshold are then pruned. 
However, these criteria-based methods often require numerous trial and error to find a suitable configuration that fits within the desired resource budget. 
Learning-based methods employ a mask function to learn scalar mask parameters associated with the significance of each channel, which are then binarized using the mask module. This approach often requires less trial and error, enhancing its efficiency in identifying the optimal configuration within the desired resource constraints. These characteristics position learning-based methods as advanced and promising approaches in filter pruning research. The mask parameters encompass parameters of the batch-normalization layer~\cite{liu2017learning, kang2020operation} and additional learnable parameters~\cite{huang2018data, luo2020autopruner, wang2020pruning}. The mask module includes an identity function~\cite{liu2017learning, huang2018data} and a sigmoid function with a constant steepness~\cite{kang2020operation, luo2020autopruner, wang2020pruning}. 

In our approach, we adopt the mask function with mask parameters in common with the previous learning-based works. However, in contrast to the previous works, we utilize a scheduled sigmoid function as the mask module. This scheduling allows us to stably learn binary masks, resulting in a more robust and effective filter pruning process.

\section{Backgrounds}
\label{sec:backgrouds}
We provide backgrounds on essential techniques to understand our framework.
\subsection{Tensor matricization}
\label{sec:Matricization}
In our work, \textit{Matricization} is used to transform the tensor of convolutional kernels into a matrix to conduct a singular value decomposition (SVD) operation.
Matricization is the process of reshaping the elements of an $D$-dimensional tensor $\textbf{X}\in\mathbb{R}^{I_1\times\cdots\times I_D}$ into a matrix~\cite{kolda2006multilinear, kolda2009tensor}. Let the ordered sets $\mathcal{R} = \{r_{1}, ..., r_{L}\}$ and $\mathcal{C} = \{c_{1},...,c_{M}\}$ be a partitioning of the modes $\mathcal{D} = \{1,...,D\}$.
The matricization function $\psi$ of an $D$-dimensional tensor $\textbf{X}\in\mathbb{R}^{I_1\times\cdots\times I_D}$ is defined as: 
\begin{equation}
\label{eqn:matricization}
\begin{aligned}
    \psi: \textbf{X}\xmapsto{\;\;\;\;\;\;\;}
     X\in\mathbb{R}^{J\times K}\text{,}\;\;\;\;\;\;\\ \text{where } J=\prod_{n\in \mathcal{R}}I_n \;\;\;  \text{and} \;\;\; K=\prod_{n\in \mathcal{C}}I_n
\end{aligned}
\end{equation}

For example, the weight tensor of a convolutional layer is represented as a 4-D tensor ($\textbf{W}\in\mathbb{R}^{ C_{\text{out}} \times C_{\text{in}} \times k \times k}$) where it is composed of kernels, and it can be unfolded into a matrix in many 
different forms. The two most common forms used in low-rank decomposition are as follows: \textcircled{\small 1} $W\in \mathbb{R}^{C_{\text{out}}\times (C_{\text{in}}k k)}$, 
\textcircled{\small 2} $W\in \mathbb{R}^{(C_{\text{out}}k)\times (C_{\text{in}}k)}$.

\subsection{CNN decomposition scheme}
\label{sec:decomposition}
To decompose a convolutional layer with $C_{\text{in}},~C_{\text{out}}$~(input/output channels) and  $k$ (kernel size), one of the following low-rank structures is used depending on the matricization form.

\textbf{Scheme 1:} When we use the first reshaping form \textcircled{\small 1} introduced in Section~\ref{sec:Matricization}, the convolutional weights can be considered as a linear layer with the shape of $C_{\text{out}} \times C_{\text{in}}k^{2}$. Then, the rank-$r$ approximation presents two linear mappings with weight shapes $C_{\text{out}} \times r$ and $r \times C_{\text{in}}k^{2}$. These linear mappings can be deployed as a sequence of two convolutional layers: $\textbf{W}_1 \in \mathbb{R}^{r \times C_{\text{in}} \times k \times k}$ and $\textbf{W}_2 \in \mathbb{R}^{C_{\text{out}} \times r \times 1 \times 1}$~\cite{li2018constrained, wen2017coordinating, xu2020trp}.\\

\textbf{Scheme 2:} When we use the second reshaping form \textcircled{\small 2} introduced in Section~\ref{sec:Matricization}, the convolutional weights can be considered as a linear layer of $C_{\text{out}}k \times C_{\text{in}}k$. For this scheme, an approximation of rank $r$ has two linear mappings with weight shapes $C_{\text{out}}k\times r$ and $r\times C_{\text{in}}k$. These can be implemented as a sequence of two convolutional layers as follows: $\textbf{W}_1 \in \mathbb{R}^{r \times C_{\text{in}} \times k \times 1}$, and $\textbf{W}_2 \in \mathbb{R}^{C_{\text{out}} \times r \times 1 \times k}$~\cite{jaderberg2014speeding, tai2015convolutional}.

We use \textbf{Scheme 1} throughout all experiments in our study based on the comparison results provided in Discussion~\ref{sec:discussion_folding}.

\subsection{Mask function for learning filter importance}
In our study, we adopt a mask function for learning filter importance as in the previous pruning works. A mask function consists of three components: mask parameters, mask module, and tensor matricization. 
As for the  mask module, 
NS~\cite{liu2017learning} and SCP~\cite{kang2020operation} use the parameters of the batch-normalization layer as mask parameters. SSS~\cite{huang2018data}, AutoPruner~\cite{luo2020autopruner}, and PFS~\cite{wang2020pruning} employ additional learnable parameters. As for the mask module, usually, one of two popular functions is used. NS and SSS adopt an identity function (i.e., $\phi(x)=x$). SCP, AutoPruner, and PFS use a sigmoid function defined as:
\begin{equation}
\label{eqn:sigmoid}
    \phi(x)=\frac{1}{1+\exp(-x)}
\end{equation}
For each weight $\textbf{W}_l$ of the $l$-th layer, mask is defined as the mask parameter $M_l$, mask module $\phi$, and tensor matricization (See Eq.~(\ref{eqn:matricization})) as below:
\begin{equation}
    \mathcal{T}(\textbf{W}_l, M_l)=\psi^{-1}\big(\phi(M_l)\cdot\psi(\textbf{W}_l)\big)
\end{equation}

\subsection{Singular Value Thresholding}
\label{sec:svt}
For differentiable threshold learning of the rank reduction,  we adopt the Singular Value Thresholding~(SVT) operator~\cite{cai2010singular}. Consider the Singular Value Decomposition (SVD) of a matrix $X \in \mathbb{R}^{m \times n}$ of rank $r$. 
\begin{equation}
\label{eqn:svd}
    X=U\Sigma V^T, \Sigma = \text{diag}(\sigma_1,\cdots,\sigma_r),
\end{equation}
where $\sigma_i$ is $i$-th singular value of $X$.
For a $\gamma \geq 0$, soft-thresholding operator SVT is defined as below:
\begin{equation}
\label{eqn:svt}
\begin{aligned}
    \text{SVT}(X, \gamma) \coloneqq U \cdot D(\Sigma,\gamma) \cdot V^T,
\end{aligned}
\end{equation}
where $U$ is an $m\times m$ real unitary matrix, $\Sigma$ is an $m\times n$ rectangular diagonal matrix with non-negative real values in the diagonal, $V$ is an $n\times n$ real unitary matrix, and $D(\Sigma,\gamma)=\text{diag}([\sigma_1-\gamma]_{+},\cdots,[\sigma_r-\gamma]_+)$.

\section{Differentiable Framework for hybrid structured compression}
\label{sec:dof}
We first formulate the CCOP to select the filters and rank.
Then, we reparameterize the discrete parameters in the CCOP to continuous parameters by employing a continuous relaxation. Additionally, we approximate the budget constraint as a penalty term by applying Lagrangian relaxation.

\subsection{CCOP of hybrid structured compression} 
\label{sec:back_formulation}
We mathematically formulate the CCOP to select the filter and rank in a network. The problem
can be formulated as a constrained combinatorial optimization problem whose objective is to determine the optimal selection $\nu$ that achieves as high performance as possible while satisfying the desired resource budget $\mathcal{B}_{d}$. The budget can be in terms of FLOPs, MACs, or the number of parameters. We formulate the optimization problem as follows:
\begin{equation}
\label{eqa:general_formula}
    \min_{\nu}\mathcal{L}\big(h(\textbf{W}, \nu)\big) \;\;
    ~\text{s.t.}\;\;~\mathcal{B}(\nu) \le \mathcal{B}_{d},
\end{equation}
where 
$\textbf{W}$ is the pre-trained weights, $\nu$ is the set of selection parameters for determining the structural modifications of $\textbf{W}$, $h$ is a function that returns weights whose structure is determined by $\nu$, $\mathcal{L}$ is the loss function (e.g., cross-entropy), $\mathcal{B(\cdot)}$ is a function that measures the required resource budget, and $\mathcal{B}_{d}$ is the desired resource budget.

For filter pruning, $\nu$ is the collection of binary channel masks associated with the output filter of each layer (i.e., $\mathbf{c}=\{c_1,\cdots,c_L\mid c_l\in\{0,1\}^{p_l}$\} where $L$ is the number of layers and $p_l$ is the number of output filters of the $l$-th layer). 
For low-rank decomposition, $\nu$ is the collection of ranks of each layer (i.e., $\mathbf{r}=\{r_1,\cdots,r_L\mid r_l\in\mathbb{N}\}$).
For hybrid compression, $\nu$ is a collection of the binary filter masks and the integer rank of each layer (i.e., $\{(c_l, r_l)\}_{l=1}^L$). 

\subsection{Differentiable optimization formulation for CCOP}
\label{sec:4.1}
Our principled goal is to transform the discrete optimization problem with a constraint in Eq.~(\ref{eqa:general_formula}) into a continuous optimization problem that can be approximately solved by gradient-based optimization in an end-to-end manner.
To achieve this, we first introduce a differentiable surrogate function over the continuous parameters from which the discrete parameters can be recovered using continuous functions. 
Additionally, we integrate the budget constraint into the objective.
More specifically, we re-define the problem~in Eq.~(\ref{eqa:general_formula}) using a continuous relaxation of the parameters and a Lagrangian relaxation of the constraint as:
\begin{equation}
\label{eqa:optimization_objective_of_nu}
\begin{aligned}
    \min_{\mathbf{\mathbf{z_c},\mathbf{z_r}}}&\mathcal{L}\big(h_{\text{DF}}(\textbf{W}, \mathbf{z_c,z_r})\big)\\
    & + \lambda\left\Vert\mathcal{B}\big(g_{\mathbf{c}}(\mathbf{z_c}), g_{\mathbf{r}}(\mathbf{z_r})\big) - \mathcal{B}_{d}\right\Vert^2
\end{aligned}
\end{equation}
where $h_{\text{DF}}$ is a differentiable surrogate function, $g_{\mathbf{c}}(\mathbf{z_c})$ is a continuous function that can recover the number of filters for each layer from the continuous parameter $\mathbf{z_c}$, $g_{\mathbf{r}}(\mathbf{z_r})$ is a continuous function that can recover the rank for each layer from the continuous parameter $\mathbf{z_r}$, and $\lambda$ is a Lagrange multiplier used to regularize the budget constraint. 
Our objective, as expressed in Eq.~(\ref{eqa:optimization_objective_of_nu}), is to optimize the continuous parameters $\mathbf{z_c}$ and $\mathbf{z_r}$. These parameters play a crucial role in approximating the optimal filter and rank across all $L$ layers while satisfying the imposed budget constraint.

\section{Methods}
\label{sec:method}
In Eq.~(\ref{eqa:optimization_objective_of_nu}), there are several components need to be carefully considered. These involve continuous parameters $\mathbf{z_c}$ and $\mathbf{z_r}$, the surrogate function $h_{\text{DF}}$, and the continuous functions $g_{\mathbf{c}}$ and $g_{\mathbf{r}}$.
In this section, we provide our implementation details.

\subsection{Differentiable Surrogate function}
\subsubsection{Differentiable Mask Learning}
\label{sec:scheduled_mask_optimization}
To construct the part of $h_{\text{DF}}$ that depends on $\mathbf{z_c}$ for the filter selection, we implement a differentiable function $\mathcal{T}_{\text{DML-S}}$ named \textit{Differentiable Mask Learning with Scheduled Sigmoid~(DML-S).}
Let $\mathbf{z_c}=\{M_1,\cdots,M_L \;|\; M_l \in \mathbb{R}^{C_{\text{out}}^l \times C_{\text{out}}^l}\}$ be a set of diagonal matrices.
We adopt a scheduled factor into the sigmoid function for generating the approximate binary masks as below:
\begin{equation}
\label{eqa:steepeness scheduled}
\begin{aligned}
    \phi_{s}(x, i) = \frac{1}{1+\exp\big(-1*\mu_{i}*(x-0.5)\big)},
\end{aligned}
\end{equation} 
where $\mu_{i} = \text{min}(\alpha, \mu_{i-1}$+$\beta$). Note that $\mu_{i}$ is the scheduled factor affecting the steepness of the sigmoid in iteration $i$. It is updated in every iteration and does not exceed $\alpha$.
$\mu_{i}$ is kept at a very low value in the early phase of training, and it is increased as the optimization process progresses. 
When $\mu_{i}$ becomes large enough, the values of approximate binary masks converge to either 0 or 1. 
That is, it is completely determined which filter should be removed. 
The hyper-parameter settings of DML-S are provided in Appendix~\ref{sec:implementation_details}.

For each weight $\textbf{W}_l$ of the $l$-th layer, we define $\mathcal{T}_{\text{DML-S}}$ using Eq.~(\ref{eqn:matricization}) and Eq.~(\ref{eqa:steepeness scheduled}) as below:
\begin{equation}
\label{eqn:s_m}
    \begin{aligned}
        \mathcal{T}_{\text{DML-S}}(\textbf{W}_l, M_l, i) = \psi^{-1}\big(\phi_{s}(M_l, i)\cdot \psi(\textbf{W}_l)\big)
    \end{aligned}
\end{equation}
Since all the underlying functions in Eq.~(\ref{eqn:s_m}) are differentiable, it can be confirmed that $\mathcal{T}_{\text{DML-S}}$ is differentiable.

As in the previous works on filter pruning~\cite{huang2018data,wang2020pruning}, we adopt the masking function, where the masks are learnable parameters. There are two justification for adopting this function. Firstly, its implementation is straightforward. Secondly, we aim to employ gradient-based optimization to solve Eq~(\ref{eqa:optimization_objective_of_nu}) in an end-to-end manner.
In contrast to the previous works, however, we adopt a \textit{scheduled} sigmoid function as a mask module because 
the scheduling technique has been proven to be useful for gradient-based optimization in the general deep learning field~\cite{kwon2020nimble,loshchilov2016sgdr, zhai2022scaling,zhou2021trar}.
Although the introduction of the scheduled factor is a simple technique, 
we find that this can result in a significant improvement in performance as shown in Table~\ref{tab:scheduled_sigmoid}.

\begin{table}[t]
\small
  \centering
  \begin{tabular}{lccc}
\toprule
Flop reduction rate                                      & 0.5            & 0.6            & 0.7   \\ \hline
Sigmoid with $\mu_{i}$          & \textbf{94.13} & \textbf{93.20} & \textbf{92.76} \\
Sigmoid without $\mu_{i}$                                         & 92.84          & 92.77          & 92.25 \\ 
\bottomrule
  \end{tabular}
  \caption{
  Performance comparison of implementing $\mathcal{T}_{\text{DML-S}}$ with and without scheduled factor $\mu_{i}$. Results are shown for ResNet56 on CIFAR10.
  }
  \label{tab:scheduled_sigmoid}
\end{table}

\subsubsection{Differentiable Threshold Learning}
\label{sec:differentiable_threshold_rank}
To formulate the component of $h_{\text{DF}}$ that depends on $\mathbf{z_r}$ for the rank selection, we construct a differentiable function $\mathcal{T}_{\text{DTL-S}}$ named \textit{Differentiable Threshold Learning with Singular value thresholding~(DTL-S).}
Let $\mathbf{z_r}=\{\gamma_1,\cdots,\gamma_L \;|\; \gamma_l \in \mathbb{R}\}$ be a set of real parameters. To construct $\mathcal{T}_{\text{DTL-S}}$, we adopt SVT operator discussed in Section~\ref{sec:svt}. 

For each weight $\textbf{W}_l$ of the $l$-th layer, we define $\mathcal{T}_{\text{DTL-S}}$ using Eq.~(\ref{eqn:matricization}) and SVT operator as below:
\begin{equation}
\label{eqn:s_t}
    \begin{aligned}
        \mathcal{T}_{\text{DTL-S}}(\textbf{W}_l,\gamma_l) = \psi^{-1}\big(\text{SVT}(\psi(\textbf{W}_l),\gamma_l)\big)
    \end{aligned}
\end{equation}
As in $\mathcal{T}_{\text{DML-S}}$, we can easily confirm that $\mathcal{T}_{\text{DTL-S}}$ is differentiable.
The previous works have utilized heuristic or iterative search techniques for optimal rank selection and stayed away from an integration of gradient-based optimization that allows an end-to-end framework.
In our work, we implement a differentiable function to learn continuous parameters associated with the rank. By constructing the DTL-S, we can jointly optimize the continuous parameters over all $L$-layers using the gradient-based optimization.

\subsubsection{The overall surrogate function $h_{\text{DF}}$}
The differentiable surrogate function is defined as the composition of $\mathcal{T}_{\text{DML-S}}$ and $\mathcal{T}_{\text{DTL-S}}$ as shown below:
\begin{equation}
\label{eqn:surrogate_ftn_conv}
    h_{\text{DF}}(\textbf{W}_l, M_l, \gamma_l, i) = \mathcal{T}_{\text{DTL-S}}(\mathcal{T}_{\text{DML-S}}(\textbf{W}_l, M_l, i),\gamma_l)
\end{equation}
$h_{\text{DF}}$ is differentiable because $\mathcal{T}_{\text{DML-S}}$ and $\mathcal{T}_{\text{DTL-S}}$ are differentiable. The overall training procedure is provided in Algorithm~\ref{alg:combined_alg}.

\begin{figure*}[t]
    \centering
    \begin{subfigure}{0.95\textwidth}
        \centering
        \includegraphics[width=\textwidth]{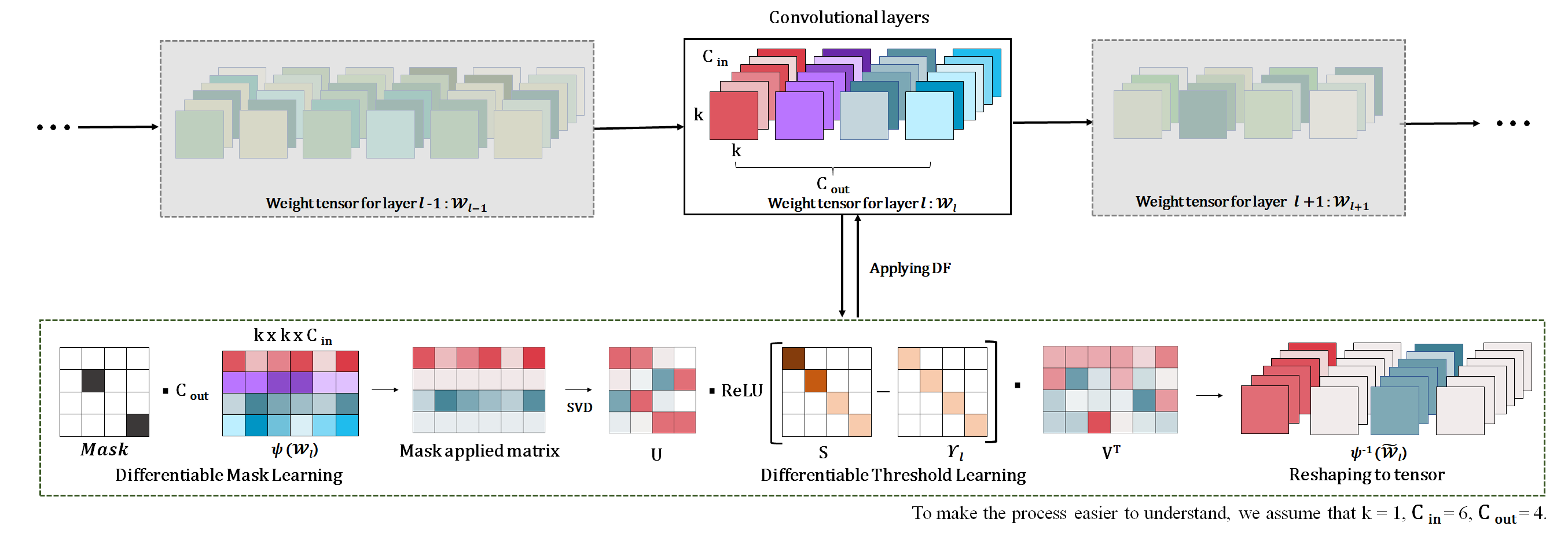}
    \end{subfigure}
    \hfill
\caption{Illustration of DF algorithm's forward process.}
\label{fig:overall_process}
\end{figure*}

Figure~\ref{fig:overall_process} shows the forward process of DF. 
The two functions of DF are simultaneously optimized using gradient-based optimization while the pre-trained network's weights are kept frozen. After the gradient-based optimization is completed, the binary mask in $\mathbf{z_c}$ and the rank in $g_\mathbf{r}(\mathbf{z_r})$ are rounded such that the compressed structure can be determined. Then, we first prune the filters whose binary mask is zero and subsequently perform low-rank decomposition of the pruned weights. Following the common convention in low-rank studies~\cite{idelbayev2020low, phan2020stable, xu2020trp}, matrix decomposition is not applied when the chosen rank is large and results in an increase in FLOPs.

\begin{algorithm}[!t]
\footnotesize
    \caption{The pseudocode of DF algorithm}
    \label{alg:combined_alg}
    \textbf{Input}: desired resource budget~$\mathcal{B}_d$;
                    model parameters $\mathbf{W} = \{\textbf{W}_{l}\}_{l=1}^{L}.$\\
    \textbf{Output}: mask variables $\mathbf{z_c} = \{M_{l}\}_{l=1}^{L}$; thresholding variables $\mathbf{z_r} = \{\gamma_{l}\}_{l=1}^{L}$\\
    \textbf{Require}: function for measuring resource budget $\mathcal{B}(\cdot,\cdot)$; epoch E; iteration T
    \begin{algorithmic}[1]
        \Statex Initialize $\mathbf{z_c}$ = \textbf{1}, $\mathbf{z_r}$ = \textbf{0}
        \Statex Calculate $\mathcal{B}_{\text{cal}}=\mathcal{B}(g_{\mathbf{c}}(\mathbf{z_c}), g_{\mathbf{r}}(\mathbf{z_r}))$
        \While{$\Vert\mathcal{B}_{\text{cal}}-\mathcal{B}_{d}\Vert \geq \epsilon$}
        \For{$\text{ep} = 1 \colon \text{E}$}
            \For{$\text{i} = 1 \colon \text{T}$}
                \For{$l = 1 \colon L$}
                    \State $\textbf{W}_l^{\text{comp}}=h_{\text{DF}}(\textbf{W}_l, M_l, \gamma_l, i)$ (Eq.~(\ref{eqn:surrogate_ftn_conv}))
                 
                \EndFor    
            \State $\mathbf{W}^{\text{comp}} = \{\textbf{W}_l^{\text{comp}}\}_{l=1}^{L}$
            \State $\mathcal{B}_{\text{cal}}=\mathcal{B}(g_{\mathbf{c}}(\mathbf{z_c}), g_{\mathbf{r}}(\mathbf{z_r}))$
      
            \State {\textbf{Update} $\mathbf{z_r}, \mathbf{z_c} \text{ as} \newline
                    \hspace*{5.5em} \arg\min_{\mathbf{z_r}, \mathbf{z_c}}\mathcal{L}(\mathbf{W}^{\text{comp}}) + \lambda\left\Vert\mathcal{B}_{\text{cal}} - \mathcal{B}_{d}\right\Vert^2$}
            \EndFor
        \EndFor
        \EndWhile
    \end{algorithmic}
\end{algorithm}

\subsection{Differentiable budget function}
\label{sec:differentiable_budget_function}
To complete our framework, it is required to represent the discrete function within the budget constraint as a differentiable function.
To achieve this goal, we first design two differentiable functions {$g_{\mathbf{c}}$ and $g_{\mathbf{r}}$ that can be used for constructing a differentiable budget function $\mathcal{B}$.
First, to approximate the number of filters corresponding to the continuous parameter $\mathbf{z_c}$, we define the set-valued function $g_{\mathbf{c}}$ as below:
\begin{equation}
\label{eqa:mask learning}
\begin{aligned}
    g_\mathbf{c}(\mathbf{z_c}) =\Big\{\textbf{1}^{T}\cdot \text{diag}\big(\phi(M_l)\big)\Big\}_{l=1}^{L}
\end{aligned}
\end{equation}
where $\phi$ is a sigmoid function~(See Eq.~(\ref{eqn:sigmoid})).
Second, to approximate the rank corresponding to the continuous parameter $\mathbf{z_r}$, we define the set-valued function $g_{\mathbf{r}}$ as below:
\begin{equation}
\label{eqa:rank_generator}
\begin{aligned}
    g_{\mathbf{r}}(\mathbf{z_r}) = \Big\{\textbf{1}^{T}\cdot\Big(\tanh\big(\text{ReLU}(\Sigma_l-\gamma_l) \cdot \tau\big)\Big)\Big\}_{l=1}^{L},\\
\end{aligned}
\end{equation}
where $\Sigma_l$ is a diagonal matrix whose diagonal entries are singular values of the $l$-th layer weight matrix $W_l$. 
As in the introduction of the sigmoid function for filter pruning, we introduce a hyperbolic tangent function with ReLU in Eq.~(\ref{eqa:rank_generator}) to determine the rank. A scaling hyper-parameter $\tau$ is used to control the steepness of $\tanh(\cdot)$.
Note that $g_\mathbf{c}(\mathbf{z_c})$ and $g_\mathbf{r}(\mathbf{z_r})$ are differentiable because their underlying functions are differentiable.

From the $g_\mathbf{c}(\mathbf{z_c})$ and $g_{\mathbf{r}}(\mathbf{z_r})$, the budget function $\mathcal{B}(g_{\mathbf{c}}(\mathbf{z_c}), g_{\mathbf{r}} (\mathbf{z_r}))$ can be defined. Following the previous works~\cite{he2018soft, hou2022chex, hua2019channel, ruan2020edp, tang2020scop}, we choose FLOP as the resource budget. Then, the FLOP resource budget can be defined as below:
\begin{equation}
\label{eqn:budget_flops}
 \frac{\sum_{l=1}^{L} A_l \cdot g_{\mathbf{r}}(\mathbf{z_r})(l) \cdot \big(k_l \cdot k_l\cdot g_{\mathbf{c}}(\mathbf{z_c})(l-1) + g_{\mathbf{c}}(\mathbf{z_c})(l)\big)}{\sum_{l=1}^{L} A_l \cdot k_l \cdot k_l \cdot C_{\text{in}}^l \cdot C_{\text{out}}^l},
\end{equation}
where $A_l$ denotes the area of the $l$-th layer's feature maps and $k_l$ is the kernel size of the $l$-th layer. $C_{\text{in}}^l$ and $C_{\text{out}}^l$ denote $l$-th layer's input and output channels of the original pre-trained weight, respectively. $g_{\mathbf{r}}(\mathbf{z_r})(l)$ and $g_{\mathbf{c}}(\mathbf{z_c})(l)$ are the $l$-th layer's selected rank and the number of selected filters, respectively. 
Because all the elements in Eq.~(\ref{eqn:budget_flops}) are differentiable, the budget function $\mathcal{B}(g_{\mathbf{c}}(\mathbf{z_c}), g_{\mathbf{r}}(\mathbf{z_r}))$ is also differentiable. Although we use FLOP as the resource budget, other types of resource budget can be formulated as well.

\section{Experiments}
\label{sec:exp}

\begin{figure*}[t]
    \centering
    \begin{subfigure}[b]{0.47\textwidth}
        \centering
        \includegraphics[width=\textwidth]{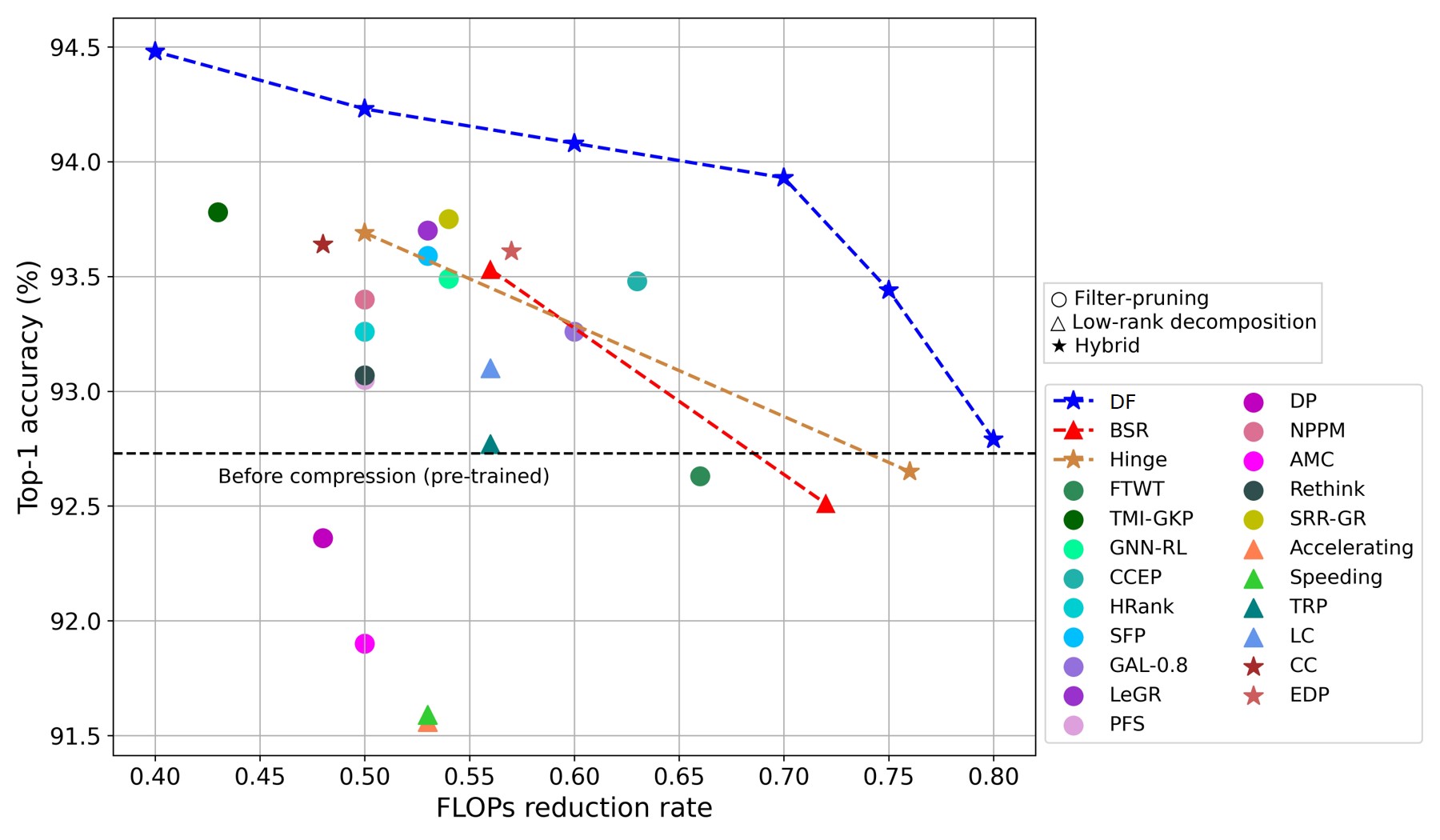}
        \caption{ResNet56 on CIFAR10}
        \label{fig:CIFAR10_resnet56}
    \end{subfigure}
    \hfill
    \begin{subfigure}[b]{0.47\textwidth}
        \centering
        \includegraphics[width=\textwidth]{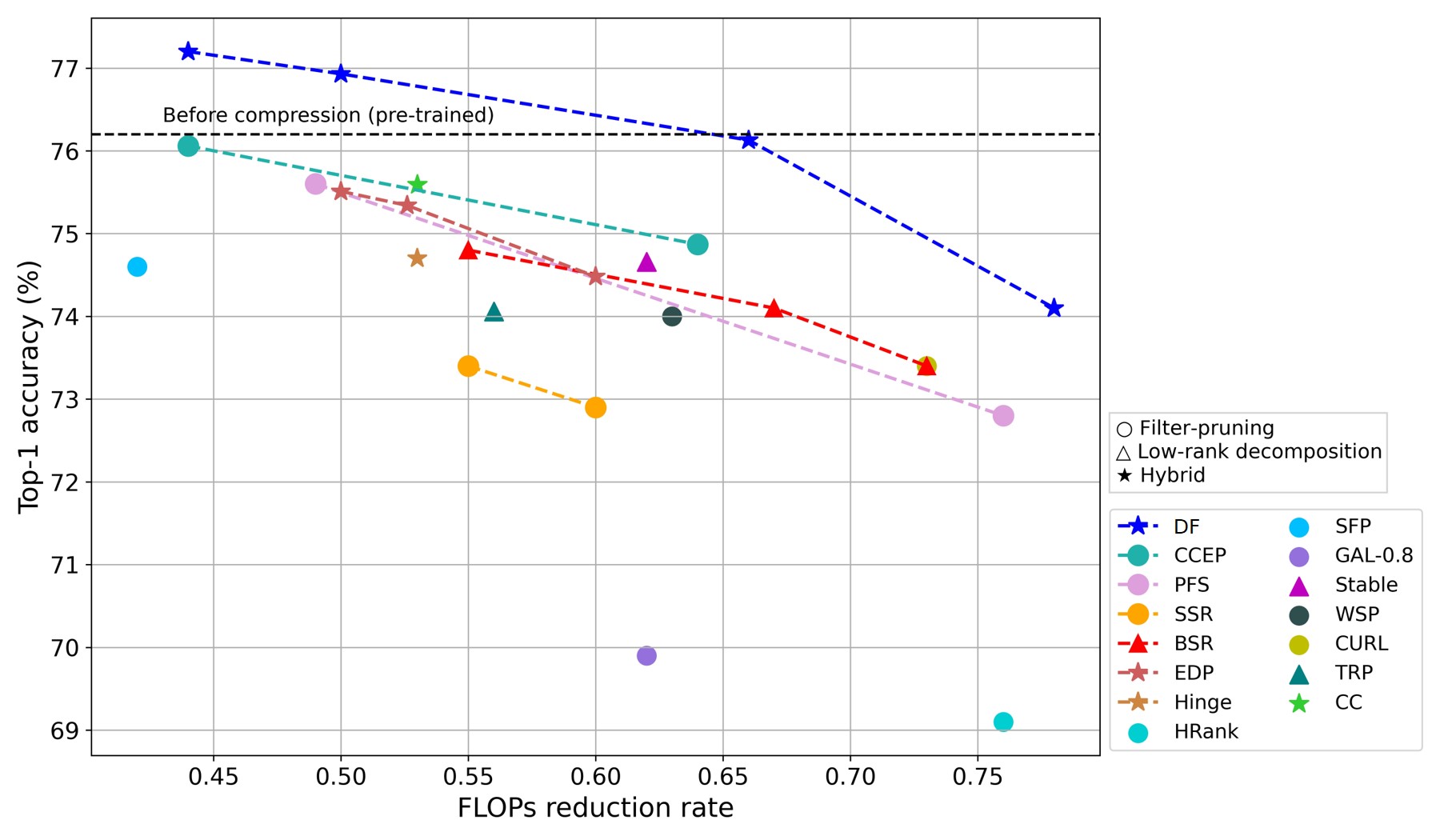}
        \caption{ResNet50 on ImageNet}
        \label{fig:ImageNet_resnet50}
    \end{subfigure}
    \hfill
\caption{Comparison of our method with SoTA pruning, low-rank decomposition, and hybrid compression methods for (a) ResNet56 on CIFAR10 and (b) ResNet50 on ImageNet.}
\label{fig:cifar10_imagenet}
\end{figure*} 
\subsection{Experimental settings}
To validate the effectiveness and generalizability of DF, we evaluate its compression performance on five benchmarks: VGG16 on CIFAR10, ResNet56 on CIFAR10 and CIFAR100, and ResNet50 and MobileNetV2 on ImageNet-1k. We follow the original VGG and ResNet training settings to prepare a baseline network. For ResNet50 and MobileNetV2, we use the official PyTorch pre-trained models. The standard data augmentation techniques are used for all the datasets. The full training dataset is used to learn masks and thresholds for the ranks.  As in the previous studies~\cite{alvarez2017compression, alwani2022decore, cai2021soft, chin2020towards, idelbayev2020low}, we fine-tune the compressed model to improve performance further. 

\subsection{Implementation details}
\label{sec:implementation_details}
For reproducibility, we provide the details of implementation and hyper-parameters used for training DF. Our implementations are based on LC~\cite{idelbayev2020low} library. Our implementation codes will be made available online. 

\subsubsection{Hyper-parameter setting}
For all the experiments, we use the hyper-parameters in Table~\ref{tap:hyperparameters}. For $\mu_i$, its $\alpha$ is simply set as a large constant of 50 because DF's performance is not sensitive to the choice, and its $\beta$ is explored using a light grid search.
\begin{table}[h]
  \centering
  \resizebox{0.95\columnwidth}{!}{\begin{tabular}{lccccccccc}
    \toprule
    Hyper-parameter                 & CIFAR10 & CIFAR100 & ImageNet  \\
    \midrule
    Batch size                      & 128     & 128      & 64        \\
    Epochs for fine-tuning          & 100     & 100      & 100        \\
    Learning rate for baseline model  & 0.01  &  0.01    & -    \\
    Learning rate for fine-tuning        & 0.001  & 0.001    & 0.001     \\
    $\lambda$        & 1 & 1   & 1    \\
    $\tau$        & $\frac{2}{\sigma_1}$ & $\frac{2}{\sigma_1}$   & $\frac{2}{\sigma_1}$    \\
    $\mu_0$      & 5 & 5   & 5    \\
    $\alpha$ (in $\mu_i$)       & 50 & 50   & 50    \\
    $\beta$ (in $\mu_i$)     & 4 & 4   & 8 \\
    \bottomrule
  \end{tabular}}
  \caption{Hyper-parameters used for training DF on various experiments. $\sigma_1$ is the largest singular value for each layer.}
  \label{tap:hyperparameters}
\end{table}

\subsubsection{Tuning details of hyper-parameters}
We fix the hyper-parameter $\lambda = 1$ in Eq.~(\ref{eqa:optimization_objective_of_nu}). Both $\tau$ in Eq.~(\ref{eqa:rank_generator}) and $\mu_i$ in Eq.~(\ref{eqa:steepeness scheduled}) are tuned via a light grid-search. 
$\tau$ controls which singular values should be treated as zero. We define $\tau$ as $\frac{C}{\sigma_1}$, where $\sigma_1$ is the largest singular value of each layer, and perform a light grid search for $C$. The normalization by $\sigma_1$ allows a single hyper-parameter $C$ to be used over all the layers. Therefore, the normalization significantly simplifies the hyper-parameter search.
For $\mu_i$, its $\alpha$ is simply set as a large constant of 50 because DF's performance is not sensitive to the choice, and its $\beta$ is explored using a light grid search.
\subsection{Experimental results}
Two of the most popular benchmarks are CIFAR-10 on ResNet56 and ImageNet-1k on ResNet50. From the existing literature, we have gathered a comprehensive collection of comparisons for these benchmarks. Detailed comparison tables are presented in Table~\ref{tab:supplementary_CIFAR10} for CIFAR-10 and Table~\ref{tab:supplementary_ImageNet} for ImageNet-1k.
also provide graphical summaries in Figure~\ref{fig:cifar10_imagenet} for the two cases where a sufficiently large number of comparisons exist. For the other benchmarks, we provide table summaries where fewer comparison points are available. As for hybrid compression, several works have been published recently. One of the most recent and meaningful works we know of is CC~\cite{li2021towards}, and it is also compared with DF.
\begin{table*}[!hbt]
\centering
\resizebox{0.95\textwidth}{!}{
\begin{tabular}{ccccccccc}
\hline
\begin{tabular}[c]{@{}c@{}} Dataset\\ on Model\end{tabular} &
\begin{tabular}[c]{@{}c@{}} Compression\\ method\end{tabular} &
Algorithm &
Baseline (\%) &
Test acc.(\%) & 
$\Delta$ Test acc.(\%) & 
\begin{tabular}[c]{@{}c@{}} MFLOPs\\ (Reduction ratio)\end{tabular}&
\begin{tabular}[c]{@{}c@{}} Params\\ (Compression ratio)\end{tabular}\\
\hline 

\multirow{33}{*}{\begin{tabular}[c]{@{}c@{}} CIFAR10\\ on ResNet56\end{tabular}} & 
\multirow{7}{*}{Low-rank} &
Accelerating~\cite{zhang2015accelerating}  & 93.14 & 91.56 & - 1.58 & 58.9 (53\%) & N/A\\
&&Speeding~\cite{jaderberg2014speeding}  & 93.14 & 91.59 & - 1.55 & 58.9 (53\%) & N/A\\
&&LC~\cite{idelbayev2020low}  & 92.73 & 93.10 & + 0.37 & 55.7 (56 \%)& N/A\\
&&TRP~\cite{xu2020trp}  & 93.14 & 92.77 & - 0.37 & 52.9 (57 \%)& N/A\\
&&CA~\cite{alvarez2017compression}  & 92.73 & 91.13 & - 1.60 & 51.4 (59 \%)& N/A\\ 
&&BSR~\cite{eo2023effective}  & 92.73 & 93.53 & + 0.80 & 55.7 (56 \%)& 0.37 M (56 \%)\\
&&BSR~\cite{eo2023effective}  & 92.73 & 92.51 & - 0.22 & 32.1 (74 \%)& 0.21 M (75 \%)\\
\cline{2-8}
&\multirow{18}{*}{Pruning} &
CHIP~\cite{sui2021chip}   & 93.26 &94.16 & + 0.75 & 66.0 (47 \%)& 0.48 M (43 \%) \\
&&DP~\cite{kim2019differentiable}  & 92.66 & 92.36 & - 0.30 & 65.2 (48 \%)& N/A\\ 
&&AMC~\cite{he2018amc}  & 92.80 & 91.90 & - 0.90 & 62.7 (50 \%)& N/A\\ 
&&CP~\cite{he2017channel}  & 93.80 & 92.80 & - 1.00 & 62.7 (50 \%)& N/A\\
&&ThiNet~\cite{luo2017thinet}  & 93.80 & 92.98 & - 0.82 & 62.7 (50 \%)& 0.41 M (50 \%)\\
&&PFS~\cite{wang2020pruning}  & 93.23 & 93.05 & - 0.18 & 62.7 (50 \%)& N/A\\ 
&&Rethink~\cite{liu2018rethinking}  & 93.80 & 93.07 & - 0.73 & 62.7 (50 \%)& N/A\\
&&SFP~\cite{he2018soft}  & 93.59 & 93.35 & - 0.24 & 62.7 (50 \%) & N/A\\
&&NPPM~\cite{gao2021network}  & 93.04 & 93.40 & + 0.36 & 62.7 (50 \%)& N/A\\
&&DCP~\cite{zhuang2018discrimination}  & 93.80 & 93.49 & - 0.31 & 62.7 (50 \%)& 0.43 M (49 \%)\\
&&LeGR~\cite{chin2020towards}  & 93.90 & 93.70 & - 0.20 & 58.9 (53 \%)& N/A \\
&&SRR-GR~\cite{wang2021convolutional}  & 93.38 & 93.75 & - 0.37 & 57.9 (54 \%)& N/A \\
&&GNN~\cite{yu2022topology}  & 93.39 & 93.49 & + 0.10 & 57.6 (54 \%)& N/A \\
&&FTWT~\cite{elkerdawy2022fire}  & 93.66 & 92.63 & - 1.03 & 47.4 (60 \%)& N/A\\
&&ASFP~\cite{he2019asymptotic}  & 94.85 & 89.72 & - 5.13 & 35.2 (73 \%)& N/A \\
&&ASRFP~\cite{cai2021softer}  & 94.85 & 90.54 & - 4.31 & 35.2 (73 \%)& N/A\\ 
&&GHFP~\cite{cai2021soft}  & 94.85 & 92.54 & - 2.31 & 35.2 (73 \%)& N/A\\
&&CHIP \cite{sui2021chip}  & 93.26 & 92.05 & - 1.21 & 34.8 (73 \%)& 0.24 M (72 \%) \\\cline{2-8}
&\multirow{8}{*}{Hybrid} &
CC~\cite{li2021towards}  & 93.33 & 93.64 & + 0.31 & 65.2 (48 \%)& 0.44 M (48 \%)\\ 
&&Hinge~\cite{li2020group}  & 92.95 & 93.69 & + 0.74 & 62.7 (50 \%)& 0.41 M (51 \%)\\
&& \textbf{DF}  & 92.73 & \textbf{94.23} & + 1.50 & 62.7 (50 \%)& 0.43 M (49 \%) \\ 
&& \textbf{DF}  & 92.73 & \textbf{94.13} & + 1.40 & 55.1 (55 \%)& 0.39 M (54 \%)\\ 
&&EDP~\cite{ruan2020edp} & 93.61 & 93.61 & 0.00 & 53.0 (58 \%)& 0.39 M (54 \%)\\
&& \textbf{DF} & 92.73 & \textbf{94.08} & + 1.35 & 47.4 (60 \%)& 0.37 M (57 \%)\\
&& \textbf{DF}  & 92.73 & \textbf{93.44} & + 0.71 &  32.2 (75 \%)& 0.22 M (74 \%)\\  
&&Hinge~\cite{li2020group}  & 92.95 & 92.65 & - 0.30 & 31.0 (76 \%)& 0.17 M (80 \%)\\\hline

\end{tabular}
}
\caption{\label{tab:supplementary_CIFAR10}
Full performance comparison results of CIFAR-10 on ResNet56.} 

\label{tab:CIFAR10_full}
\end{table*}

\begin{table*}[!hbt]
\centering
\resizebox{0.95\textwidth}{!}{%
\begin{tabular}{ccccccccc}
\hline

\begin{tabular}[c]{@{}c@{}} Dataset\\ on Model\end{tabular} &
\begin{tabular}[c]{@{}c@{}} Compression\\ method\end{tabular} &
Algorithm & 
Baseline (\%) &
Test acc.(\%) & 
$\Delta$ Test acc.(\%) & 
\begin{tabular}[c]{@{}c@{}} GFLOPs\\ (Reduction ratio)\end{tabular}& 
\begin{tabular}[c]{@{}c@{}} Params\\ (Compression ratio)\end{tabular} \\\hline

\multirow{32}{*}{\begin{tabular}[c]{@{}c@{}} ImageNet-1k\\ on ResNet50\end{tabular}} &\multirow{6}{*}{Low-rank} &
TRP \cite{xu2020trp} & 75.90 & 74.06 & - 1.84 & 2.3 (45 \%)& N/A \\
&& Stable~\cite{phan2020stable} &76.15 & 74.66 & - 1.47 & 1.6 (62 \%)& N/A\\
&& BSR~\cite{eo2023effective} &76.20 & 75.00 & - 1.2 & 2.2 (47 \%)& 12.5 M (51 \%)\\
&& BSR~\cite{eo2023effective} &76.20 & 74.80 & - 1.4 & 1.8 (55 \%)& 10.1 M (60 \%)\\
&& BSR~\cite{eo2023effective} &76.20 & 74.10 & - 2.1 & 1.4 (67 \%)& 7.5 M (70 \%)\\
&& BSR~\cite{eo2023effective} &76.20 & 73.40 & - 2.8 & 1.1 (73 \%)& 5.0 M (80 \%)\\
\cline{2-8} 

&\multirow{19}{*}{Pruning} 
& PFS~\cite{wang2020pruning}  & 76.10 & 76.70 & + 0.60 & 3.0 (25 \%)& N/A\\
&&ThiNet~\cite{luo2017thinet}  & 73.00 & 72.04 & - 0.96 & 2.4 (37 \%)& 16.9 M (34 \%)\\ 
&& SFP~\cite{he2018soft}  &76.20 & 74.60 & - 1.60 & 2.4 (42 \%)& N/A\\ 
&&CCEP~\cite{shang2022neural}  & 76.13 & 76.06  & - 0.07 & 2.3 (44 \%)& N/A \\ 
&&CHIP \cite{sui2021chip}  & 76.20 & 76.30 & + 0.10 & 2.3 (45 \%) & 15.1 M (41 \%)\\ 
&&PFS~\cite{wang2020pruning}  &76.10 & 75.60 & - 0.50 & 2.0 (49 \%)& N/A \\ 
&&CP~\cite{he2017channel}  &76.10 & 73.30 & - 2.80 & 2.1 (49 \%)& N/A\\
&&CP~\cite{he2017channel}  &76.10 & 73.30 & - 2.80 & 2.1 (51 \%)& N/A\\
&& SSR~\cite{lin2019toward}  &76.20 & 73.40 & - 2.80 & 1.9 (55 \%)& 15.5 M (39 \%)\\
&& SSR~\cite{lin2019toward}  &76.20 & 72.60 & - 3.60 & 1.7 (60 \%)& 12.0 M (53 \%)\\ 
&& GAL~\cite{lin2019towards}  &76.20 & 69.90 & - 6.30 & 1.6 (62 \%)&14.6 M (43 \%)\\
&& WSP~\cite{guo2021weak}  &76.13 & 73.91 & - 2.22 & 1.5 (63 \%)& 11.6 M (54 \%)\\ 
&& CCEP~\cite{shang2022neural}  & 76.13 & 74.87  & - 1.26 & 1.5 (64 \%) & N/A\\ 
&& AutoPruner~\cite{luo2020autopruner}  &76.10 & 73.00 & - 3.10 & 1.4 (65 \%)& N/A\\ 
&& WSP~\cite{guo2021weak}  &76.13 & 72.04 & - 4.09 & 1.1 (73 \%)& 9.1 M (65 \%)\\ 
&& CURL~\cite{luo2020neural}  &76.20 & 73.40 & - 2.80 & 1.1 (73 \%)& 6.7 M (74 \%)\\
&& PFS~\cite{wang2020pruning}  &76.10 & 72.80 & - 3.30 & 1.0 (76 \%)& N/A\\ 
&& CHIP \cite{sui2021chip}   &76.20 & 73.30 & - 2.90 & 1.0 (76 \%)& 7.8 M (69 \%)\\
&& HRANK~\cite{lin2020hrank}   &76.10 & 69.10 & - 7.00 & 1.0 (76 \%)&8.3 M (67 \%)\\ \cline{2-8} 
&\multirow{10}{*}{Hybrid} 
&EDP~\cite{ruan2020edp} &75.90 &75.51  & - 0.39 & 2.1 (50 \%) & N/A\\ 
&& \textbf{DF} &76.20 & \textbf{77.20}  & + 1.00 & 2.3 (44 \%) & 13.8 M (46 \%)\\ 
&& \textbf{DF} &76.20 & \textbf{76.93}  & + 0.73 & 2.1 (50 \%) & 12.2 M (52 \%)\\ 
&&EDP~\cite{ruan2020edp} &75.90 & 75.34 & - 0.56 & 1.9 (53 \%)& N/A\\ 
&&CC~\cite{li2021towards}  &76.15 & 75.59  & - 0.56 & 1.9 (53 \%) & 13.2 M (48 \%)\\ 
&&Hinge~\cite{li2020group}  & N/A & 74.7 &  N/A & 1.9 (53 \%)& N/A\\
&&EDP~\cite{ruan2020edp} &75.90 &74.48  & - 1.42 & 1.7 (60 \%)& N/A\\ 
&&CC~\cite{li2021towards} &76.15 & 74.54  & - 1.61 & 1.5 (63 \%) & 10.6 M (59 \%)\\ 
&& \textbf{DF}  &76.20 & \textbf{76.13}  & - 0.07 & 1.4 (66 \%)& 8.4 M (67 \%)\\ 
&& \textbf{DF} &76.20 & \textbf{74.10}  & - 2.10 & 0.9 (78 \%)& 5.4 M (79 \%)\\\hline

\end{tabular}
}
\caption{\label{tab:supplementary_ImageNet}
Full performance comparison results of ImageNet-1k on ResNet50.} 
\label{tab:imagenet_full}
\end{table*}
\textbf{ResNet56 on CIFAR10}
Figure~\ref{fig:CIFAR10_resnet56} shows the comparison results for ResNet56 on CIFAR10. DF outperforms the existing methods by a large margin across all FLOP reduction rates. In particular, the 50\% FLOP reduction rate has been investigated by a large number of existing methods, and DF achieves the best performance under the same constraint. Note that DF consistently exhibits a higher performance than the uncompressed baseline model across all the FLOP reduction rates. In particular, DF can reduce the FLOPs by 40\% while improving the accuracy by 1.6\% when compared to the uncompressed baseline model. 

\begin{table*}[!t]
\centering
\resizebox{0.95\textwidth}{!}{
\begin{tabular}{c|c|l|lcccccc}
\hline
Dataset &
Model &
\begin{tabular}[c]{@{}c@{}} Compression\\ method\end{tabular} &
Algorithm & 
Baseline (\%) &
Test acc.(\%) & 
$\Delta$ Test acc.(\%) & 
\begin{tabular}[c]{@{}c@{}} MFLOPs\\ (Reduction rate)\end{tabular}&
\begin{tabular}[c]{@{}c@{}} Params\\ (Compression rate)\end{tabular}\\
\hline \hline
\multirow{9}{*}{CIFAR10}&\multirow{10}{*}{VGG16}&\multirow{1}{*}{Low-rank}& LC~\cite{idelbayev2020low}  & 93.43 & 93.83 & + 0.40 & 132 (57.8 \%)&3.16 M (78.7 \%)\\ \cline{3-9}
&&\multirow{4}{*}{Pruning} 
&SSS~\cite{huang2018data} & 93.96 & 93.02 & - 0.94 & 183 (41.6 \%)&3.93 M (73.8 \%) \\ 
&&&CP~\cite{he2017channel} & 93.99 & 93.67 & - 0.32 & 156 (50.0 \%)& N/A\\
&&&GAL-0.1~\cite{lin2019towards} & 93.96 & 93.42 & - 0.54 & 141 (54.8 \%)&12.25 M (17.8 \%)\\
&&&DECORE~\cite{alwani2022decore}  & 93.96 & 93.56 & - 0.40 & 110 (64.8 \%) & 1.66 M (89.0 \%)\\\cline{3-9}
&&\multirow{4}{*}{Hybrid}& Hinge~\cite{li2020group}  & 94.02 & 93.59 & - 0.43 & 122 (60.9 \%)&11.92 M (20.0 \%)\\ 
&&&\textbf{DF}  & 94.14 & \textbf{93.87} & - 0.27 & 125 (60.0 \%) & 2.24 M (85.0 \%)\\ 
&&&\textbf{DF}  & 94.14 & \textbf{93.74} & - 0.40 & 109 (65.0 \%) & 1.79 M (88.0 \%)\\
&&&\textbf{DF}  & 94.14 & \textbf{93.64} & - 0.50 & 94 (70.0 \%) &1.47 M (90.1 \%)\\\hline\hline
\multirow{8}{*}{CIFAR100}&\multirow{8}{*}{ResNet56}&\multirow{2}{*}{Low-rank} &CA~\cite{alvarez2017compression} & 72.39 & 64.79 & - 7.60 & 75 (40.0 \%)& N/A\\
&&&LC~\cite{idelbayev2020low}  & 72.39 & 69.82 & - 2.57 & 59 (52.6 \%)& N/A\\\cline{3-9}
&&\multirow{4}{*}{Pruning}&ASFP~\cite{he2019asymptotic} & 72.92 & 69.35 & - 3.57 & 59 (52.6 \%)& N/A \\
&&&ASRFP~\cite{cai2021softer} & 72.92 & 69.16 & - 3.32 & 59 (52.6 \%)& N/A\\ 
&&&GHFP~\cite{cai2021soft}  & 72.92 & 69.62 & - 3.30 & 59 (52.6 \%)& N/A\\
&&&PGMPF~\cite{cai2022prior}  & 72.92 & 70.21 & - 2.71 & 59 (52.6 \%)& N/A\\ \cline{3-9}
&&\multirow{2}{*}{Hybrid}& \textbf{DF} & 72.39 & \textbf{72.12} & - 0.27 & 63 (50.0 \%) & 0.41 M (51.7 \%) \\
&&&\textbf{DF} & 72.39 & \textbf{71.05} & - 1.34 & 55 (55.0 \%)
& 0.37 M (56.4 \%)\\   \hline 
\end{tabular}}
\caption{Performance comparison results of VGG16 on CIFAR10 and ResNet56 on CIFAR100.} 
\label{tab:cifar}
\end{table*}

\textbf{VGG16 on CIFAR10}
Table~\ref{tab:cifar} presents the comparison results for VGG16 on CIFAR10.
The performance of DF is superior to the filter pruning, low-rank decomposition, and hybrid methods that were compared; in particular, DF shows no noticeable drop in performance even at higher FLOP reduction rates (maximum of 0.5 percentage point is dropped). The performance of DF at 70\% reduction rate is superior to the performance of Hinge~\cite{li2020group} at 60\% reduction rate.

\textbf{ResNet56 on CIFAR100}
The results for ResNet56 on CIFAR100 can be also found in Table~\ref{tab:cifar}. DF is able to preserve the original model's performance with only a small loss even when the model is compressed by 50\% or more. Compared to the other methods, DF exhibits the best performance even when its FLOPs is reduced by a large percentage (@~55\% reduction).

\textbf{ResNet50 and ResNet18 on ImageNet}
The results for ResNet50 on ImageNet can be founded in Figure~\ref{fig:ImageNet_resnet50}. The graphical summary confirms that DF achieves a superior performance over all FLOP reduction rates. For instance, when we compare the difference in the FLOP rate between our method and the CC algorithm~\cite{li2021towards} at the same performance (75.59\%), our method can accelerate the inference time by 14\% more than the CC method~\cite{li2021towards} (0.53 \textit{vs.} 0.68). In addition, even when ResNet50 is compressed by 50\% FLOP reduction, our method exhibits higher performance than the uncompressed baseline performance.

The results for ResNet18 on ImageNet are summarized in Table~\ref{tab:all_dataset}. No experimental results of hybrid algorithms were available for ResNet18 on ImageNet. When compared with the low-rank and filter pruning methods, DF outperforms them in all FLOP reduction rates. Even when the model is compressed by 70\%, the performance is still better than the uncompressed baseline model. 

\textbf{MobileNetV2 on ImageNet}
Performance comparison results for ImageNet on light-weight MobileNetV2 are summarized in Table~\ref{tab:all_dataset}. MobileNetV2 is a well-known computationally efficient
model, which makes it harder to compress. Nevertheless, our method can increase the model's top-1 accuracy to 72.16\% from the uncompressed baseline's 71.80\% while reducing the FLOPs by 35\%. Furthermore, the performance reduction is only 0.17 percentage point when the FLOP reduction rate is 55\%. 
\begin{table*}[!t]
\centering
\resizebox{0.95\textwidth}{!}{
\begin{tabular}{c|c|l|lccccc}
\hline
Dataset &
Model &
\begin{tabular}[c]{@{}c@{}} Compression\\ method\end{tabular} &
Algorithm & 
Baseline (\%) &
Test acc.(\%) & 
$\Delta$ Test acc.(\%) & 
\begin{tabular}[c]{@{}c@{}} GFLOPs\\ (Reduction rate)\end{tabular} &
\begin{tabular}[c]{@{}c@{}} Params\\ (Compression rate)\end{tabular} \\
\hline \hline

\multirow{26}{*}{ImageNet}&\multirow{17}{*}{ResNet18}&\multirow{3}{*}{Low-rank}

&Stable~\cite{phan2020stable}  &69.76 & 68.62 & - 1.14 & 1.00 (45 \%) & N/A \\
&&& TRP~\cite{xu2020trp}  &69.10 & 65.51 & - 3.59 & 0.73 (60 \%) & N/A \\
&&& ALDS~\cite{liebenwein2021compressing}  &69.62 & 69.24 & - 0.38 & 0.64 (65 \%) & N/A \\\cline{3-9}
&&\multirow{11}{*}{Pruning}&SFP~\cite{he2018soft}  &70.28 & 67.10 & - 3.18 & 1.06  (42 \%) & N/A\\ 
&&& FPGM~\cite{he2019filter} & 70.28 & 68.41 & - 1.87 & 1.06  (42 \%) & 7.10 M (39 \%)\\
&&& PFP~\cite{liebenwein2019provable} & 69.74 & 65.65  & - 4.09  & 1.04 (43 \%) & N/A\\
&&& DMCP~\cite{guo2020dmcp}  &N/A & 69.00 & \;\;N/A & 1.04 (43 \%) & N/A  \\
&&& CHEX~\cite{hou2022chex}  &N/A & 69.60 & \;\;N/A & 1.03 (43 \%) & N/A  \\
&&& SCOP~\cite{tang2020scop}  &69.76 & 68.62 & - 1.14 & 1.00 (45 \%) & N/A  \\
&&& FBS~\cite{gao2018dynamic}  &69.76 & 68.17 & - 1.59 & 0.91 (50 \%) & N/A  \\
&&& CGNET~\cite{hua2019channel}  &69.76 & 68.30 & - 1.46 & 0.89 (51 \%) & N/A  \\
&&& GNN~\cite{yu2022topology}  &69.76 & 68.66 & -1.10 & 0.89 (51 \%) & N/A  \\
&&& ManiDP~\cite{tang2021manifold}  &69.76 & 68.88 & - 0.88 & 0.89 (51 \%) & N/A  \\
&&& PGMPF~\cite{cai2022prior}  &70.23 & 66.67 & - 3.56 & 0.84 (54 \%)& N/A \\\cline{3-9}
&&\multirow{3}{*}{Hybrid}
&\textbf{DF} & 69.76 & \textbf{71.24} & + 1.48 & 0.91 (50 \%) & 4.68 M (60 \%)\\ 
&&& \textbf{DF}  &69.76 & \textbf{70.82}  & + 1.06  & 0.70 (62 \%) &3.51 M (70 \%)\\
&&& \textbf{DF}  &69.76 & \textbf{70.15}  & + 0.39 & 0.55 (70 \%) &2.81 M (76 \%)\\\cline{2-9}
&\multirow{10}{*}{MobileNetV2}&\multirow{1}{*}{Low-rank}
& LC~\cite{idelbayev2020low}  &71.80 & 69.80 & - 2.00 & 0.21 (30 \%) & N/A \\\cline{3-9}
&&\multirow{6}{*}{Pruning}& PFS~\cite{wang2020pruning}  &71.80 & 70.90 & - 0.90 & 0.21 (30 \%) & 2.60 M (26 \%) \\
&&&AMC~\cite{he2018amc} & 71.80 & 70.80 & - 1.00 & 0.22 (27 \%) & 2.30 M (34 \%)\\
&&& MetaPruning~\cite{liu2019metapruning}  &72.00 & 71.20 & - 0.80 & 0.22 (27 \%) & N/A \\
&&& LeGR~\cite{chin2020towards}  &71.80 & 71.40 & - 0.20 & 0.21 (30 \%) & N/A\\
&&& NPPM~\cite{gao2021network} &72.02 & 72.04  & + 0.02  & 0.21 (30 \%) & N/A\\

&&& GNN~\cite{yu2022topology}  &71.87 & 70.04 & - 1.83 & 0.17 (42 \%) & N/A\\\cline{3-9}
&&\multirow{3}{*}{Hybrid}
&EDP~\cite{ruan2020edp} & N/A & 71.00 & \;\;N/A & 0.22 (27 \%) & N/A\\ \cline{4-9}
&&& \textbf{DF}  &71.80 & \textbf{72.16}  & + 0.20  & 0.19 (35 \%) & 2.24 M (36 \%)\\
&&& \textbf{DF}  &71.80 & \textbf{71.63}  & - 0.17 & 0.14 (55 \%) & 1.54 M (56 \%)\\\hline
\end{tabular}}
\caption{Performance comparison results of ResNet18 and MobileNetV2 on ImageNet.} 
\label{tab:all_dataset}
\end{table*}

\section{Discussion}
\subsection{Analysis of DML-S and DTL-S in DF}
\label{sec:6.1}

\begin{figure}[t]
    \centering
    \begin{subfigure}{0.235\textwidth}
        \centering
        \includegraphics[width=\textwidth]{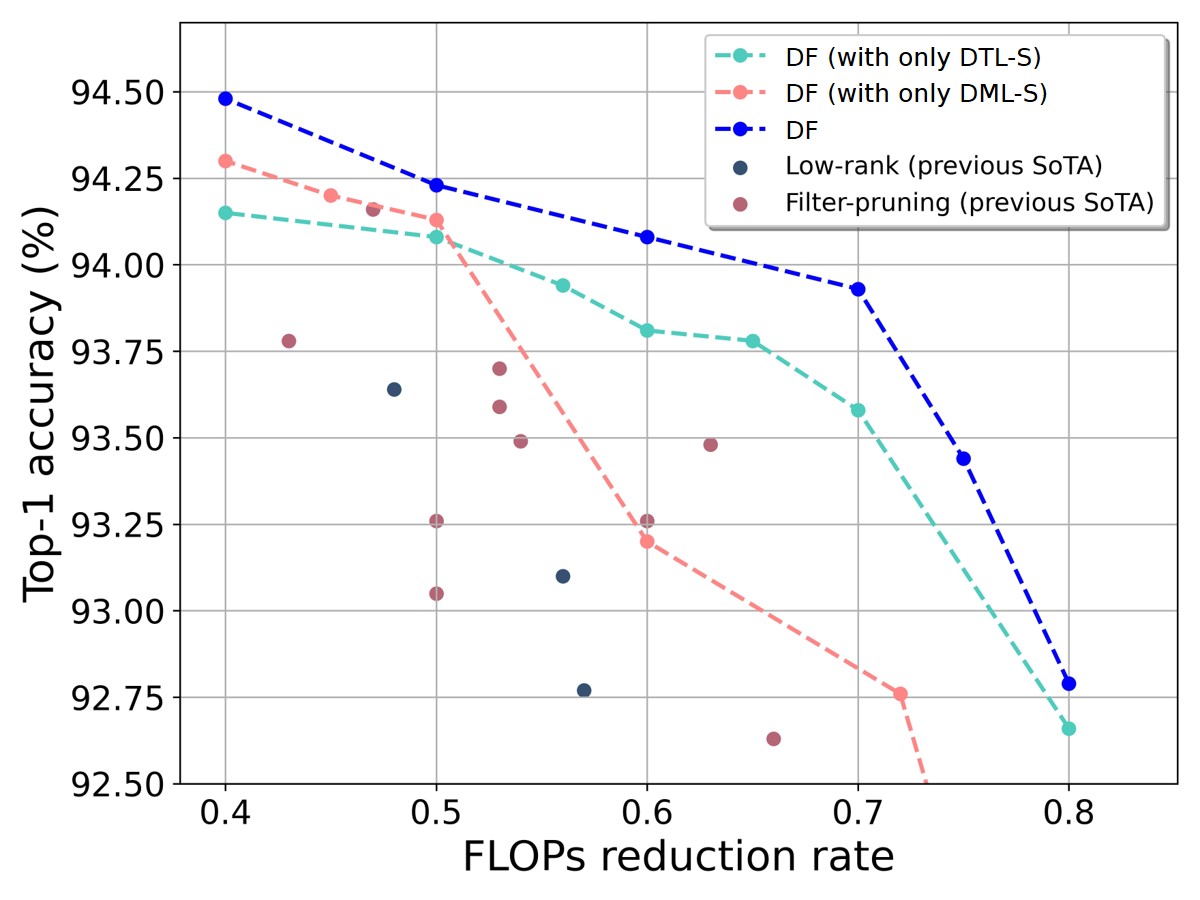}
        \caption{Performance comparison of each method.}
        \label{subfig:analysis}
    \end{subfigure}
    \hfill
    \begin{subfigure}{0.235\textwidth}
        \centering
        \includegraphics[width=\textwidth]{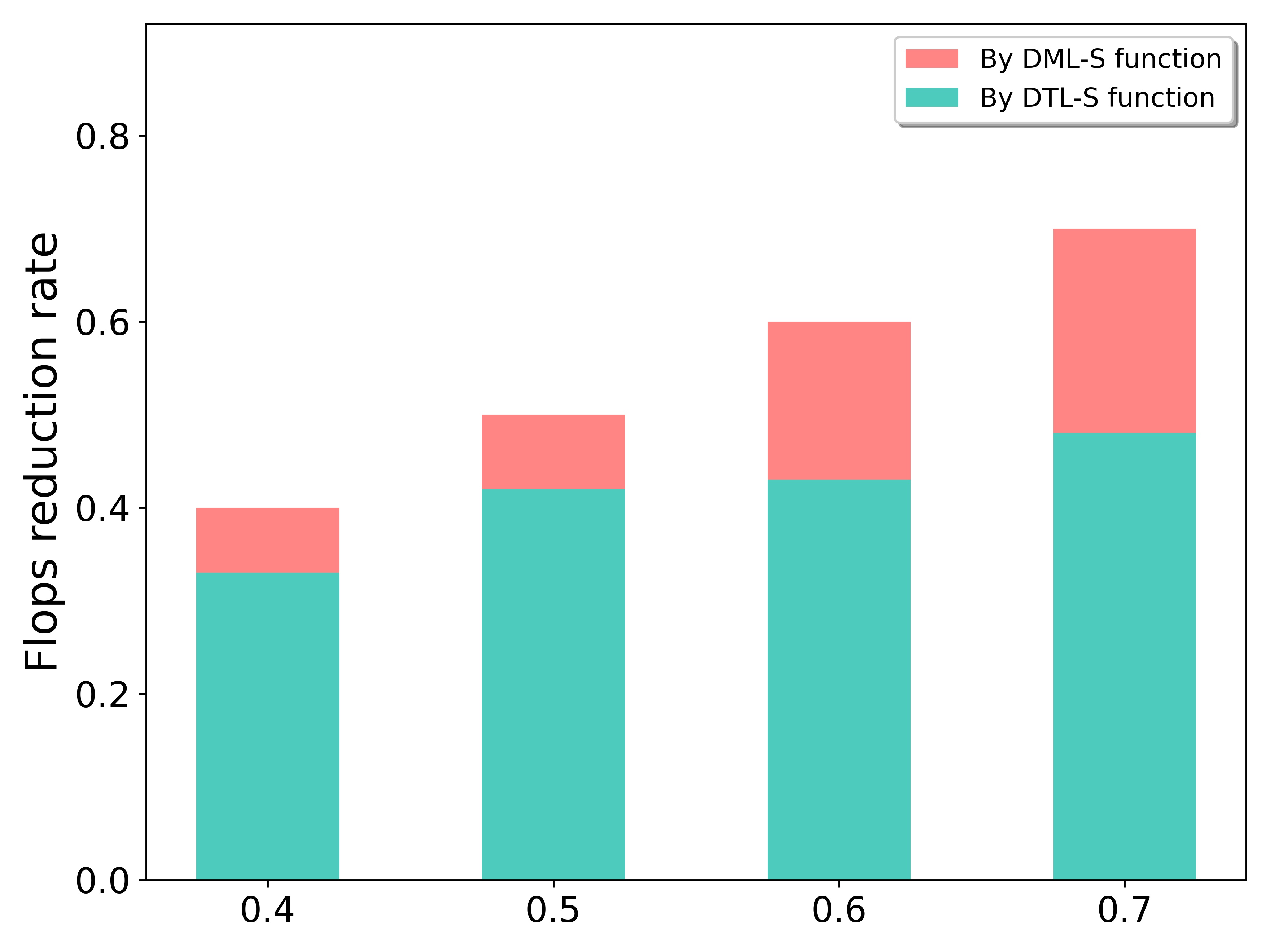}
        \caption{The contribution of each method to FLOP reduction.}
        \label{subfig:each method impact on flops reduction}
    \end{subfigure}
    \hfill
\caption{Analysis of DML-S and DTL-S: ResNet56 on CIFAR10.}

\label{fig:analysis}
\end{figure} 

DF consists of two functions (i.e., DML-S and DTL-S), and they are jointly optimized for the final filter and rank selection. For analyzing each function's effectiveness in DF, we intentionally make each function disabled in turn. The experiment is conducted for ResNet56 on CIFAR10. As shown in Figure~\ref{subfig:analysis}, the individual function works fairly well, better or comparable to most of the recent filter pruning and low-rank decomposition algorithms. In particular, DF with only DML-S~(i.e., filter pruning) shows a higher performance at relatively low FLOP reduction rates, whereas DF with only DTL-S~(i.e., low-rank decomposition) shows a higher performance at relatively high FLOP reduction rates. 
The result indicates that some of the filters are redundant, but only up to a certain level. Once the redundant filters are removed, pruning starts to suffer with a sharp performance reduction because some of the required filters need to be removed as a whole. Note that pruning does not allow a partial reduction for each filter. 
On the other hand, low-rank decomposition does not suffer from this problem because it allows a finer grain control of dimension reduction. 
As the result of jointly solving filter and rank selection, our hybrid compression algorithm DF always shows the best performance in all FLOP intervals. 

We also investigate the FLOP reduction rate by each function in DF, and the results are shown in Figure~\ref{subfig:each method impact on flops reduction}. The utilization of low-rank decomposition is substantial across all FLOP reduction rates, and the contribution of filter pruning to compression increases with the FLOP reduction rate. 

\subsection{Influence of CNN decomposition scheme}
The two matricization methods described in Section~\ref{sec:decomposition} are used in many compression studies, especially for low-rank decomposition that requires a reshaping of tensors into matrices.
Most studies use Scheme 1 rather than Scheme 2. We conducted experiments to compare which scheme performs better for our method. The results are shown in Figure~\ref{fig:scheme}. Scheme 1 demonstrates a significantly better performance, and the performance gap increases with the FLOP reduction rate. Note that typically Scheme 2 forms a matrix that is square or closer to a square matrix than Scheme 1. Therefore, the superior performance of Scheme 1 might be due to other factors such as the inherent structure of the networks that we have experimented.
\label{sec:discussion_folding}
\begin{figure}[t]
    \centering
    \begin{subfigure}{0.225\textwidth}
        \centering
        \includegraphics[width=\textwidth]{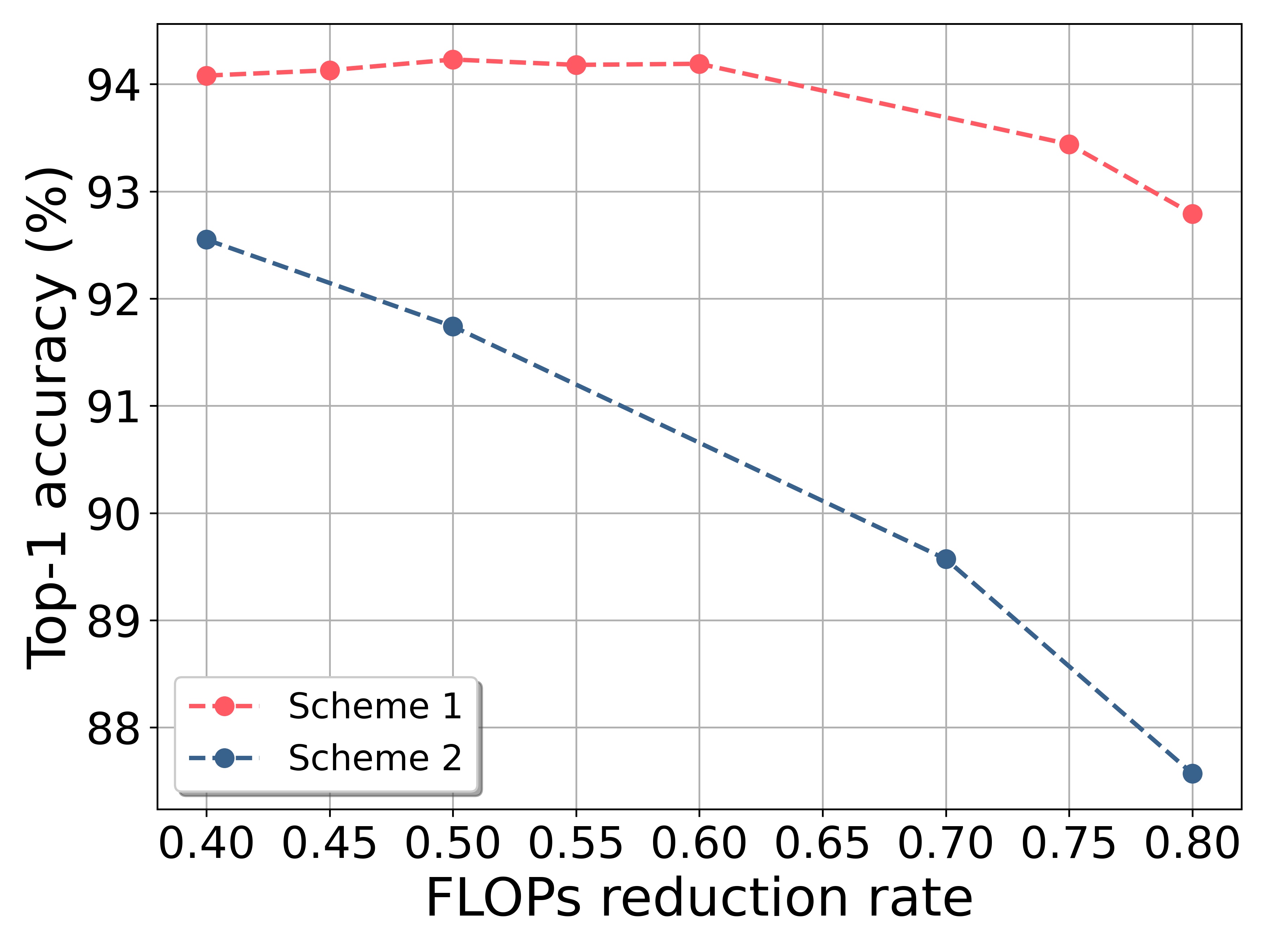}
        \caption{ResNet56 on CIFAR10}
    \end{subfigure}
    \hfill
    \begin{subfigure}{0.225\textwidth}
        \centering
        \includegraphics[width=\textwidth]{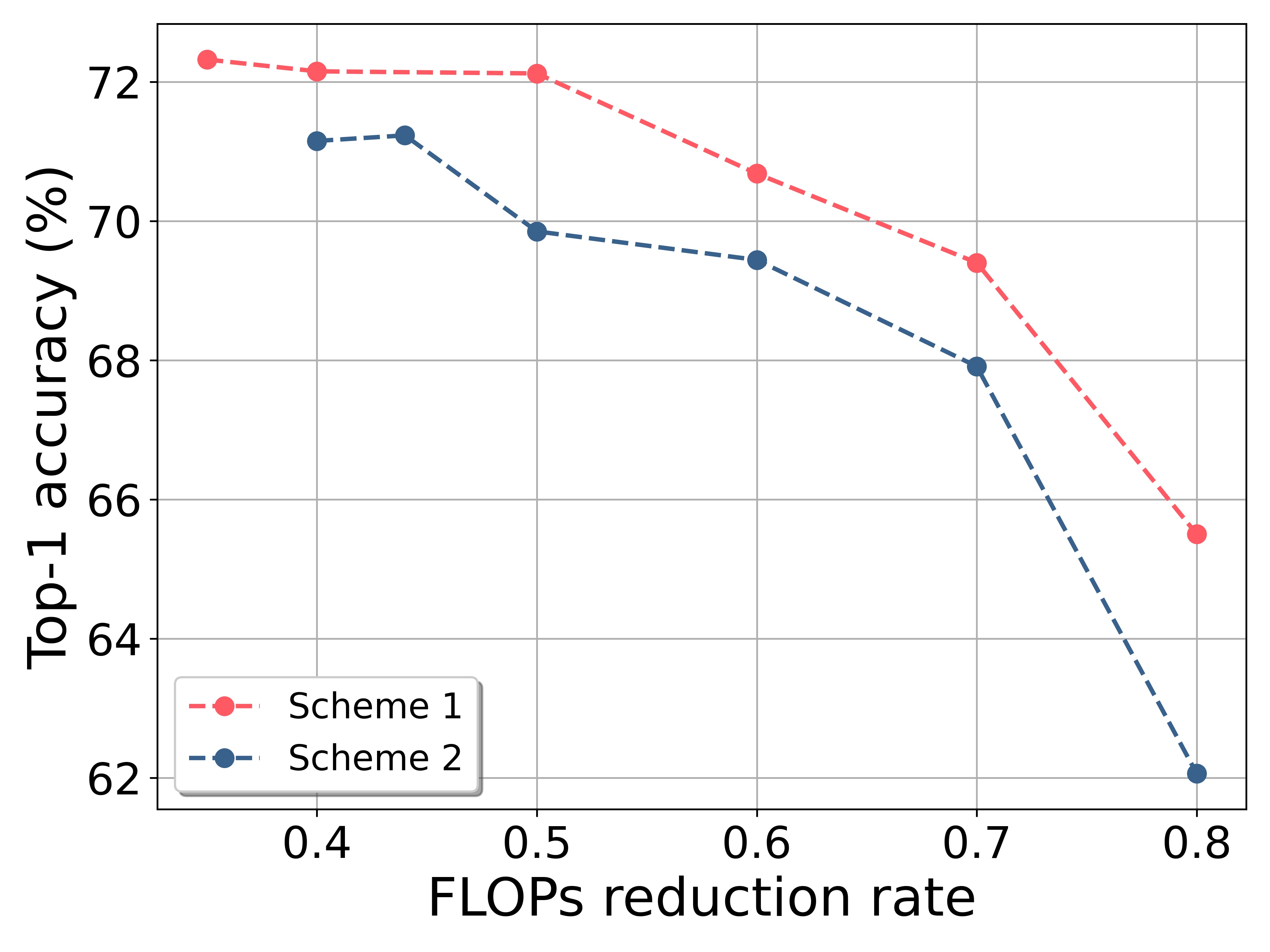}
        \caption{ResNet56 on CIFAR100}
    \end{subfigure}
    \hfill
\caption{Comparison of matricization schemes.}
\label{fig:scheme}
\end{figure}

\section{Conclusion}
\label{sec:Concolusion}

In this investigation, we have introduced a novel differentiable optimization framework, DF, tailored for hybrid structured compression. Through the integration of two distinct differentiable learning strategies, our framework has been shown to be highly effective: differentiable mask learning employing scheduled sigmoid for filter pruning, and differentiable threshold learning employing singular value thresholding for low-rank decomposition. These strategies operate cohesively, achieving joint optimization while adhering to the specified resource constraint.

Importantly, our differentiable optimization framework obviates the necessity for compression-aware regularization during training, thereby eliminating a computational overhead. Empirical results validate the efficiency and effectiveness of our compression method, particularly in the context of vision tasks.
While this study primarily focused on vision tasks, the potential applicability of our approach to domains such as natural language processing and audio tasks, employing large-scale networks like transformers, represents an intriguing avenue for future exploration. Our research lays a foundation for broader advancements in structured compression techniques.




\bibliographystyle{unsrtnat}
\bibliography{reference}

\begin{thebibliography}{72}
\providecommand{\natexlab}[1]{#1}
\providecommand{\url}[1]{\texttt{#1}}
\expandafter\ifx\csname urlstyle\endcsname\relax
  \providecommand{\doi}[1]{doi: #1}\else
  \providecommand{\doi}{doi: \begingroup \urlstyle{rm}\Url}\fi

\bibitem[Idelbayev and Carreira-Perpin{\'a}n(2020)]{idelbayev2020low}
Yerlan Idelbayev and Miguel~A Carreira-Perpin{\'a}n.
\newblock Low-rank compression of neural nets: Learning the rank of each layer.
\newblock In \emph{Proceedings of the IEEE/CVF Conference on Computer Vision
  and Pattern Recognition}, pages 8049--8059, 2020.

\bibitem[Li et~al.(2020)Li, Gu, Mayer, Gool, and Timofte]{li2020group}
Yawei Li, Shuhang Gu, Christoph Mayer, Luc~Van Gool, and Radu Timofte.
\newblock Group sparsity: The hinge between filter pruning and decomposition
  for network compression.
\newblock In \emph{Proceedings of the IEEE/CVF conference on computer vision
  and pattern recognition}, pages 8018--8027, 2020.

\bibitem[Ruan et~al.(2020)Ruan, Liu, Yuan, Li, Hu, Li, and
  Maybank]{ruan2020edp}
Xiaofeng Ruan, Yufan Liu, Chunfeng Yuan, Bing Li, Weiming Hu, Yangxi Li, and
  Stephen Maybank.
\newblock Edp: An efficient decomposition and pruning scheme for convolutional
  neural network compression.
\newblock \emph{IEEE Transactions on Neural Networks and Learning Systems},
  32\penalty0 (10):\penalty0 4499--4513, 2020.

\bibitem[Shang et~al.(2022)Shang, Wu, Hong, and Qian]{shang2022neural}
Haopu Shang, Jia-Liang Wu, Wenjing Hong, and Chao Qian.
\newblock Neural network pruning by cooperative coevolution.
\newblock \emph{arXiv preprint arXiv:2204.05639}, 2022.

\bibitem[Sui et~al.(2021)Sui, Yin, Xie, Phan, Aliari~Zonouz, and
  Yuan]{sui2021chip}
Yang Sui, Miao Yin, Yi~Xie, Huy Phan, Saman Aliari~Zonouz, and Bo~Yuan.
\newblock Chip: Channel independence-based pruning for compact neural networks.
\newblock \emph{Advances in Neural Information Processing Systems},
  34:\penalty0 24604--24616, 2021.

\bibitem[Zhuang et~al.(2018)Zhuang, Tan, Zhuang, Liu, Guo, Wu, Huang, and
  Zhu]{zhuang2018discrimination}
Zhuangwei Zhuang, Mingkui Tan, Bohan Zhuang, Jing Liu, Yong Guo, Qingyao Wu,
  Junzhou Huang, and Jinhui Zhu.
\newblock Discrimination-aware channel pruning for deep neural networks.
\newblock \emph{arXiv preprint arXiv:1810.11809}, 2018.

\bibitem[Han et~al.(2016)Han, Liu, Mao, Pu, Pedram, Horowitz, and
  Dally]{han2016eie}
Song Han, Xingyu Liu, Huizi Mao, Jing Pu, Ardavan Pedram, Mark~A Horowitz, and
  William~J Dally.
\newblock Eie: Efficient inference engine on compressed deep neural network.
\newblock \emph{ACM SIGARCH Computer Architecture News}, 44\penalty0
  (3):\penalty0 243--254, 2016.

\bibitem[Park et~al.(2016)Park, Li, Wen, Tang, Li, Chen, and
  Dubey]{park2016faster}
Jongsoo Park, Sheng Li, Wei Wen, Ping Tak~Peter Tang, Hai Li, Yiran Chen, and
  Pradeep Dubey.
\newblock Faster cnns with direct sparse convolutions and guided pruning.
\newblock \emph{arXiv preprint arXiv:1608.01409}, 2016.

\bibitem[He et~al.(2018{\natexlab{a}})He, Kang, Dong, Fu, and Yang]{he2018soft}
Yang He, Guoliang Kang, Xuanyi Dong, Yanwei Fu, and Yi~Yang.
\newblock Soft filter pruning for accelerating deep convolutional neural
  networks.
\newblock \emph{arXiv preprint arXiv:1808.06866}, 2018{\natexlab{a}}.

\bibitem[Huang and Wang(2018)]{huang2018data}
Zehao Huang and Naiyan Wang.
\newblock Data-driven sparse structure selection for deep neural networks.
\newblock In \emph{Proceedings of the European conference on computer vision
  (ECCV)}, pages 304--320, 2018.

\bibitem[Liu et~al.(2019)Liu, Mu, Zhang, Guo, Yang, Cheng, and
  Sun]{liu2019metapruning}
Zechun Liu, Haoyuan Mu, Xiangyu Zhang, Zichao Guo, Xin Yang, Kwang-Ting Cheng,
  and Jian Sun.
\newblock Metapruning: Meta learning for automatic neural network channel
  pruning.
\newblock In \emph{Proceedings of the IEEE/CVF international conference on
  computer vision}, pages 3296--3305, 2019.

\bibitem[Luo and Wu(2020{\natexlab{a}})]{luo2020autopruner}
Jian-Hao Luo and Jianxin Wu.
\newblock Autopruner: An end-to-end trainable filter pruning method for
  efficient deep model inference.
\newblock \emph{Pattern Recognition}, 107:\penalty0 107461, 2020{\natexlab{a}}.

\bibitem[Kim et~al.(2019{\natexlab{a}})Kim, Khan, and Kyung]{kim2019efficient}
Hyeji Kim, Muhammad Umar~Karim Khan, and Chong-Min Kyung.
\newblock Efficient neural network compression.
\newblock In \emph{Proceedings of the IEEE/CVF Conference on Computer Vision
  and Pattern Recognition}, pages 12569--12577, 2019{\natexlab{a}}.

\bibitem[Liebenwein et~al.(2021)Liebenwein, Maalouf, Feldman, and
  Rus]{liebenwein2021compressing}
Lucas Liebenwein, Alaa Maalouf, Dan Feldman, and Daniela Rus.
\newblock Compressing neural networks: Towards determining the optimal
  layer-wise decomposition.
\newblock \emph{Advances in Neural Information Processing Systems},
  34:\penalty0 5328--5344, 2021.

\bibitem[Yaguchi et~al.(2019)Yaguchi, Suzuki, Nitta, Sakata, and
  Tanizawa]{yaguchi2019decomposable}
Atsushi Yaguchi, Taiji Suzuki, Shuhei Nitta, Yukinobu Sakata, and Akiyuki
  Tanizawa.
\newblock Decomposable-net: Scalable low-rank compression for neural networks.
\newblock \emph{arXiv preprint arXiv:1910.13141}, 2019.

\bibitem[Cook et~al.(1995)Cook, Lov{\'a}sz, Seymour,
  et~al.]{cook1995combinatorial}
William Cook, L{\'a}szl{\'o} Lov{\'a}sz, Paul~D Seymour, et~al.
\newblock \emph{Combinatorial optimization: papers from the DIMACS Special
  Year}, volume~20.
\newblock American Mathematical Soc., 1995.

\bibitem[Yu et~al.(2017)Yu, Liu, Wang, and Tao]{yu2017compressing}
Xiyu Yu, Tongliang Liu, Xinchao Wang, and Dacheng Tao.
\newblock On compressing deep models by low rank and sparse decomposition.
\newblock In \emph{Proceedings of the IEEE conference on computer vision and
  pattern recognition}, pages 7370--7379, 2017.

\bibitem[Li et~al.(2021)Li, Lin, Liu, Ye, Wang, Chao, Yang, Ma, Tian, and
  Ji]{li2021towards}
Yuchao Li, Shaohui Lin, Jianzhuang Liu, Qixiang Ye, Mengdi Wang, Fei Chao, Fan
  Yang, Jincheng Ma, Qi~Tian, and Rongrong Ji.
\newblock Towards compact cnns via collaborative compression.
\newblock In \emph{Proceedings of the IEEE/CVF Conference on Computer Vision
  and Pattern Recognition}, pages 6438--6447, 2021.

\bibitem[Dubey et~al.(2018)Dubey, Chatterjee, and Ahuja]{dubey2018coreset}
Abhimanyu Dubey, Moitreya Chatterjee, and Narendra Ahuja.
\newblock Coreset-based neural network compression.
\newblock In \emph{Proceedings of the European Conference on Computer Vision
  (ECCV)}, pages 454--470, 2018.

\bibitem[Guo et~al.(2019)Guo, Xie, Xu, and Xing]{guo2019compressing}
Kailing Guo, Xiaona Xie, Xiangmin Xu, and Xiaofen Xing.
\newblock Compressing by learning in a low-rank and sparse decomposition form.
\newblock \emph{IEEE Access}, 7:\penalty0 150823--150832, 2019.

\bibitem[Chen et~al.(2020)Chen, Chen, Lin, Liu, and Li]{chen2020deep}
Zhen Chen, Zhibo Chen, Jianxin Lin, Sen Liu, and Weiping Li.
\newblock Deep neural network acceleration based on low-rank approximated
  channel pruning.
\newblock \emph{IEEE Transactions on Circuits and Systems I: Regular Papers},
  67\penalty0 (4):\penalty0 1232--1244, 2020.

\bibitem[Chen et~al.(2019)Chen, Lin, Liu, Chen, Li, Zhao, and
  Yan]{chen2019exploiting}
Zhen Chen, Jianxin Lin, Sen Liu, Zhibo Chen, Weiping Li, Jin Zhao, and Wei Yan.
\newblock Exploiting weight-level sparsity in channel pruning with low-rank
  approximation.
\newblock In \emph{2019 IEEE International Symposium on Circuits and Systems
  (ISCAS)}, pages 1--5. IEEE, 2019.

\bibitem[Li et~al.(2022)Li, Pan, Chen, Ding, Zhao, and Xu]{li2022heuristic}
Nannan Li, Yu~Pan, Yaran Chen, Zixiang Ding, Dongbin Zhao, and Zenglin Xu.
\newblock Heuristic rank selection with progressively searching tensor ring
  network.
\newblock \emph{Complex \& Intelligent Systems}, 8\penalty0 (2):\penalty0
  771--785, 2022.

\bibitem[Kim et~al.(2015)Kim, Park, Yoo, Choi, Yang, and
  Shin]{kim2015compression}
Yong-Deok Kim, Eunhyeok Park, Sungjoo Yoo, Taelim Choi, Lu~Yang, and Dongjun
  Shin.
\newblock Compression of deep convolutional neural networks for fast and low
  power mobile applications.
\newblock \emph{arXiv preprint arXiv:1511.06530}, 2015.

\bibitem[Wen et~al.(2017)Wen, Xu, Wu, Wang, Chen, and Li]{wen2017coordinating}
Wei Wen, Cong Xu, Chunpeng Wu, Yandan Wang, Yiran Chen, and Hai Li.
\newblock Coordinating filters for faster deep neural networks.
\newblock In \emph{Proceedings of the IEEE International Conference on Computer
  Vision}, pages 658--666, 2017.

\bibitem[Xu et~al.(2020)Xu, Li, Zhang, Wen, Wang, Qi, Chen, Lin, and
  Xiong]{xu2020trp}
Yuhui Xu, Yuxi Li, Shuai Zhang, Wei Wen, Botao Wang, Yingyong Qi, Yiran Chen,
  Weiyao Lin, and Hongkai Xiong.
\newblock Trp: Trained rank pruning for efficient deep neural networks.
\newblock \emph{arXiv preprint arXiv:2004.14566}, 2020.

\bibitem[Zhang et~al.(2015)Zhang, Zou, He, and Sun]{zhang2015accelerating}
Xiangyu Zhang, Jianhua Zou, Kaiming He, and Jian Sun.
\newblock Accelerating very deep convolutional networks for classification and
  detection.
\newblock \emph{IEEE transactions on pattern analysis and machine
  intelligence}, 38\penalty0 (10):\penalty0 1943--1955, 2015.

\bibitem[Alvarez and Salzmann(2017)]{alvarez2017compression}
Jose~M Alvarez and Mathieu Salzmann.
\newblock Compression-aware training of deep networks.
\newblock \emph{Advances in neural information processing systems},
  30:\penalty0 856--867, 2017.

\bibitem[Li and Shi(2018)]{li2018constrained}
Chong Li and CJ~Shi.
\newblock Constrained optimization based low-rank approximation of deep neural
  networks.
\newblock In \emph{Proceedings of the European Conference on Computer Vision
  (ECCV)}, pages 732--747, 2018.

\bibitem[Li et~al.(2016)Li, Kadav, Durdanovic, Samet, and Graf]{li2016pruning}
Hao Li, Asim Kadav, Igor Durdanovic, Hanan Samet, and Hans~Peter Graf.
\newblock Pruning filters for efficient convnets.
\newblock \emph{arXiv preprint arXiv:1608.08710}, 2016.

\bibitem[He et~al.(2019{\natexlab{a}})He, Liu, Wang, Hu, and
  Yang]{he2019filter}
Yang He, Ping Liu, Ziwei Wang, Zhilan Hu, and Yi~Yang.
\newblock Filter pruning via geometric median for deep convolutional neural
  networks acceleration.
\newblock In \emph{Proceedings of the IEEE/CVF conference on computer vision
  and pattern recognition}, pages 4340--4349, 2019{\natexlab{a}}.

\bibitem[Luo et~al.(2017)Luo, Wu, and Lin]{luo2017thinet}
Jian-Hao Luo, Jianxin Wu, and Weiyao Lin.
\newblock Thinet: A filter level pruning method for deep neural network
  compression.
\newblock In \emph{Proceedings of the IEEE international conference on computer
  vision}, pages 5058--5066, 2017.

\bibitem[He et~al.(2017)He, Zhang, and Sun]{he2017channel}
Yihui He, Xiangyu Zhang, and Jian Sun.
\newblock Channel pruning for accelerating very deep neural networks.
\newblock In \emph{Proceedings of the IEEE international conference on computer
  vision}, pages 1389--1397, 2017.

\bibitem[Liu et~al.(2017)Liu, Li, Shen, Huang, Yan, and Zhang]{liu2017learning}
Zhuang Liu, Jianguo Li, Zhiqiang Shen, Gao Huang, Shoumeng Yan, and Changshui
  Zhang.
\newblock Learning efficient convolutional networks through network slimming.
\newblock In \emph{Proceedings of the IEEE international conference on computer
  vision}, pages 2736--2744, 2017.

\bibitem[Kang and Han(2020)]{kang2020operation}
Minsoo Kang and Bohyung Han.
\newblock Operation-aware soft channel pruning using differentiable masks.
\newblock In \emph{International Conference on Machine Learning}, pages
  5122--5131. PMLR, 2020.

\bibitem[Wang et~al.(2020)Wang, Zhang, Xie, Zhou, Su, Zhang, and
  Hu]{wang2020pruning}
Yulong Wang, Xiaolu Zhang, Lingxi Xie, Jun Zhou, Hang Su, Bo~Zhang, and Xiaolin
  Hu.
\newblock Pruning from scratch.
\newblock In \emph{Proceedings of the AAAI Conference on Artificial
  Intelligence}, volume~34, pages 12273--12280, 2020.

\bibitem[Kolda(2006)]{kolda2006multilinear}
Tamara~Gibson Kolda.
\newblock Multilinear operators for higher-order decompositions.
\newblock Technical report, Sandia National Laboratories (SNL), Albuquerque,
  NM, and Livermore, CA~…, 2006.

\bibitem[Kolda and Bader(2009)]{kolda2009tensor}
Tamara~G Kolda and Brett~W Bader.
\newblock Tensor decompositions and applications.
\newblock \emph{SIAM review}, 51\penalty0 (3):\penalty0 455--500, 2009.

\bibitem[Jaderberg et~al.(2014)Jaderberg, Vedaldi, and
  Zisserman]{jaderberg2014speeding}
Max Jaderberg, Andrea Vedaldi, and Andrew Zisserman.
\newblock Speeding up convolutional neural networks with low rank expansions.
\newblock \emph{arXiv preprint arXiv:1405.3866}, 2014.

\bibitem[Tai et~al.(2015)Tai, Xiao, Zhang, Wang, et~al.]{tai2015convolutional}
Cheng Tai, Tong Xiao, Yi~Zhang, Xiaogang Wang, et~al.
\newblock Convolutional neural networks with low-rank regularization.
\newblock \emph{arXiv preprint arXiv:1511.06067}, 2015.

\bibitem[Cai et~al.(2010)Cai, Cand{\`e}s, and Shen]{cai2010singular}
Jian-Feng Cai, Emmanuel~J Cand{\`e}s, and Zuowei Shen.
\newblock A singular value thresholding algorithm for matrix completion.
\newblock \emph{SIAM Journal on optimization}, 20\penalty0 (4):\penalty0
  1956--1982, 2010.

\bibitem[Kwon et~al.(2020)Kwon, Yu, Jeong, and Chun]{kwon2020nimble}
Woosuk Kwon, Gyeong-In Yu, Eunji Jeong, and Byung-Gon Chun.
\newblock Nimble: Lightweight and parallel gpu task scheduling for deep
  learning.
\newblock \emph{Advances in Neural Information Processing Systems},
  33:\penalty0 8343--8354, 2020.

\bibitem[Loshchilov and Hutter(2016)]{loshchilov2016sgdr}
Ilya Loshchilov and Frank Hutter.
\newblock Sgdr: Stochastic gradient descent with warm restarts.
\newblock \emph{arXiv preprint arXiv:1608.03983}, 2016.

\bibitem[Zhai et~al.(2022)Zhai, Kolesnikov, Houlsby, and
  Beyer]{zhai2022scaling}
Xiaohua Zhai, Alexander Kolesnikov, Neil Houlsby, and Lucas Beyer.
\newblock Scaling vision transformers.
\newblock In \emph{Proceedings of the IEEE/CVF Conference on Computer Vision
  and Pattern Recognition}, pages 12104--12113, 2022.

\bibitem[Zhou et~al.(2021)Zhou, Ren, Zhu, Sun, Liu, Ding, Xu, and
  Ji]{zhou2021trar}
Yiyi Zhou, Tianhe Ren, Chaoyang Zhu, Xiaoshuai Sun, Jianzhuang Liu, Xinghao
  Ding, Mingliang Xu, and Rongrong Ji.
\newblock Trar: Routing the attention spans in transformer for visual question
  answering.
\newblock In \emph{Proceedings of the IEEE/CVF International Conference on
  Computer Vision}, pages 2074--2084, 2021.

\bibitem[Phan et~al.(2020)Phan, Sobolev, Sozykin, Ermilov, Gusak,
  Tichavsk{\`y}, Glukhov, Oseledets, and Cichocki]{phan2020stable}
Anh-Huy Phan, Konstantin Sobolev, Konstantin Sozykin, Dmitry Ermilov, Julia
  Gusak, Petr Tichavsk{\`y}, Valeriy Glukhov, Ivan Oseledets, and Andrzej
  Cichocki.
\newblock Stable low-rank tensor decomposition for compression of convolutional
  neural network.
\newblock In \emph{European Conference on Computer Vision}, pages 522--539.
  Springer, 2020.

\bibitem[Hou et~al.(2022)Hou, Qin, Sun, Ma, Yuan, Xu, Chen, Jin, Xie, and
  Kung]{hou2022chex}
Zejiang Hou, Minghai Qin, Fei Sun, Xiaolong Ma, Kun Yuan, Yi~Xu, Yen-Kuang
  Chen, Rong Jin, Yuan Xie, and Sun-Yuan Kung.
\newblock Chex: Channel exploration for cnn model compression.
\newblock In \emph{Proceedings of the IEEE/CVF Conference on Computer Vision
  and Pattern Recognition}, pages 12287--12298, 2022.

\bibitem[Hua et~al.(2019)Hua, Zhou, De~Sa, Zhang, and Suh]{hua2019channel}
Weizhe Hua, Yuan Zhou, Christopher~M De~Sa, Zhiru Zhang, and G~Edward Suh.
\newblock Channel gating neural networks.
\newblock \emph{Advances in Neural Information Processing Systems}, 32, 2019.

\bibitem[Tang et~al.(2020)Tang, Wang, Xu, Tao, Xu, Xu, and Xu]{tang2020scop}
Yehui Tang, Yunhe Wang, Yixing Xu, Dacheng Tao, Chunjing Xu, Chao Xu, and Chang
  Xu.
\newblock Scop: Scientific control for reliable neural network pruning.
\newblock \emph{Advances in Neural Information Processing Systems},
  33:\penalty0 10936--10947, 2020.

\bibitem[Alwani et~al.(2022)Alwani, Wang, and Madhavan]{alwani2022decore}
Manoj Alwani, Yang Wang, and Vashisht Madhavan.
\newblock Decore: Deep compression with reinforcement learning.
\newblock In \emph{Proceedings of the IEEE/CVF Conference on Computer Vision
  and Pattern Recognition}, pages 12349--12359, 2022.

\bibitem[Cai et~al.(2021{\natexlab{a}})Cai, An, Yang, and Xu]{cai2021soft}
Linhang Cai, Zhulin An, Chuanguang Yang, and Yongjun Xu.
\newblock Soft and hard filter pruning via dimension reduction.
\newblock In \emph{2021 International Joint Conference on Neural Networks
  (IJCNN)}, pages 1--8. IEEE, 2021{\natexlab{a}}.

\bibitem[Chin et~al.(2020)Chin, Ding, Zhang, and Marculescu]{chin2020towards}
Ting-Wu Chin, Ruizhou Ding, Cha Zhang, and Diana Marculescu.
\newblock Towards efficient model compression via learned global ranking.
\newblock In \emph{Proceedings of the IEEE/CVF Conference on Computer Vision
  and Pattern Recognition}, pages 1518--1528, 2020.

\bibitem[Eo et~al.(2023)Eo, Kang, and Rhee]{eo2023effective}
Moonjung Eo, Suhyun Kang, and Wonjong Rhee.
\newblock An effective low-rank compression with a joint rank selection
  followed by a compression-friendly training.
\newblock \emph{Neural Networks}, 2023.

\bibitem[Kim et~al.(2019{\natexlab{b}})Kim, Park, Jung, and
  Choe]{kim2019differentiable}
Jaedeok Kim, Chiyoun Park, H~Jung, and Yoonsuck Choe.
\newblock Differentiable pruning method for neural networks.
\newblock \emph{CoRR}, 2019{\natexlab{b}}.

\bibitem[He et~al.(2018{\natexlab{b}})He, Lin, Liu, Wang, Li, and
  Han]{he2018amc}
Yihui He, Ji~Lin, Zhijian Liu, Hanrui Wang, Li-Jia Li, and Song Han.
\newblock Amc: Automl for model compression and acceleration on mobile devices.
\newblock In \emph{Proceedings of the European Conference on Computer Vision
  (ECCV)}, pages 784--800, 2018{\natexlab{b}}.

\bibitem[Liu et~al.(2018)Liu, Sun, Zhou, Huang, and Darrell]{liu2018rethinking}
Zhuang Liu, Mingjie Sun, Tinghui Zhou, Gao Huang, and Trevor Darrell.
\newblock Rethinking the value of network pruning.
\newblock \emph{arXiv preprint arXiv:1810.05270}, 2018.

\bibitem[Gao et~al.(2021)Gao, Huang, Cai, and Huang]{gao2021network}
Shangqian Gao, Feihu Huang, Weidong Cai, and Heng Huang.
\newblock Network pruning via performance maximization.
\newblock In \emph{Proceedings of the IEEE/CVF Conference on Computer Vision
  and Pattern Recognition}, pages 9270--9280, 2021.

\bibitem[Wang et~al.(2021)Wang, Li, and Wang]{wang2021convolutional}
Zi~Wang, Chengcheng Li, and Xiangyang Wang.
\newblock Convolutional neural network pruning with structural redundancy
  reduction.
\newblock In \emph{Proceedings of the IEEE/CVF Conference on Computer Vision
  and Pattern Recognition}, pages 14913--14922, 2021.

\bibitem[Yu et~al.(2022)Yu, Mazaheri, and Jannesari]{yu2022topology}
Sixing Yu, Arya Mazaheri, and Ali Jannesari.
\newblock Topology-aware network pruning using multi-stage graph embedding and
  reinforcement learning.
\newblock In \emph{International Conference on Machine Learning}, pages
  25656--25667. PMLR, 2022.

\bibitem[Elkerdawy et~al.(2022)Elkerdawy, Elhoushi, Zhang, and
  Ray]{elkerdawy2022fire}
Sara Elkerdawy, Mostafa Elhoushi, Hong Zhang, and Nilanjan Ray.
\newblock Fire together wire together: A dynamic pruning approach with
  self-supervised mask prediction.
\newblock In \emph{Proceedings of the IEEE/CVF Conference on Computer Vision
  and Pattern Recognition}, pages 12454--12463, 2022.

\bibitem[He et~al.(2019{\natexlab{b}})He, Dong, Kang, Fu, Yan, and
  Yang]{he2019asymptotic}
Yang He, Xuanyi Dong, Guoliang Kang, Yanwei Fu, Chenggang Yan, and Yi~Yang.
\newblock Asymptotic soft filter pruning for deep convolutional neural
  networks.
\newblock \emph{IEEE transactions on cybernetics}, 50\penalty0 (8):\penalty0
  3594--3604, 2019{\natexlab{b}}.

\bibitem[Cai et~al.(2021{\natexlab{b}})Cai, An, Yang, and Xu]{cai2021softer}
Linhang Cai, Zhulin An, Chuanguang Yang, and Yongjun Xu.
\newblock Softer pruning, incremental regularization.
\newblock In \emph{2020 25th International Conference on Pattern Recognition
  (ICPR)}, pages 224--230. IEEE, 2021{\natexlab{b}}.

\bibitem[Lin et~al.(2019{\natexlab{a}})Lin, Ji, Li, Deng, and
  Li]{lin2019toward}
Shaohui Lin, Rongrong Ji, Yuchao Li, Cheng Deng, and Xuelong Li.
\newblock Toward compact convnets via structure-sparsity regularized filter
  pruning.
\newblock \emph{IEEE transactions on neural networks and learning systems},
  31\penalty0 (2):\penalty0 574--588, 2019{\natexlab{a}}.

\bibitem[Lin et~al.(2019{\natexlab{b}})Lin, Ji, Yan, Zhang, Cao, Ye, Huang, and
  Doermann]{lin2019towards}
Shaohui Lin, Rongrong Ji, Chenqian Yan, Baochang Zhang, Liujuan Cao, Qixiang
  Ye, Feiyue Huang, and David Doermann.
\newblock Towards optimal structured cnn pruning via generative adversarial
  learning.
\newblock In \emph{Proceedings of the IEEE/CVF Conference on Computer Vision
  and Pattern Recognition}, pages 2790--2799, 2019{\natexlab{b}}.

\bibitem[Guo et~al.(2021)Guo, Wu, Kittler, and Feng]{guo2021weak}
Qingbei Guo, Xiao-Jun Wu, Josef Kittler, and Zhiquan Feng.
\newblock Weak sub-network pruning for strong and efficient neural networks.
\newblock \emph{Neural Networks}, 144:\penalty0 614--626, 2021.

\bibitem[Luo and Wu(2020{\natexlab{b}})]{luo2020neural}
Jian-Hao Luo and Jianxin Wu.
\newblock Neural network pruning with residual-connections and limited-data.
\newblock In \emph{Proceedings of the IEEE/CVF Conference on Computer Vision
  and Pattern Recognition}, pages 1458--1467, 2020{\natexlab{b}}.

\bibitem[Lin et~al.(2020)Lin, Ji, Wang, Zhang, Zhang, Tian, and
  Shao]{lin2020hrank}
Mingbao Lin, Rongrong Ji, Yan Wang, Yichen Zhang, Baochang Zhang, Yonghong
  Tian, and Ling Shao.
\newblock Hrank: Filter pruning using high-rank feature map.
\newblock In \emph{Proceedings of the IEEE/CVF conference on computer vision
  and pattern recognition}, pages 1529--1538, 2020.

\bibitem[Cai et~al.(2022)Cai, An, Yang, Yan, and Xu]{cai2022prior}
Linhang Cai, Zhulin An, Chuanguang Yang, Yangchun Yan, and Yongjun Xu.
\newblock Prior gradient mask guided pruning-aware fine-tuning.
\newblock In \emph{Proceedings of the AAAI Conference on Artificial
  Intelligence}, volume~1, 2022.

\bibitem[Liebenwein et~al.(2019)Liebenwein, Baykal, Lang, Feldman, and
  Rus]{liebenwein2019provable}
Lucas Liebenwein, Cenk Baykal, Harry Lang, Dan Feldman, and Daniela Rus.
\newblock Provable filter pruning for efficient neural networks.
\newblock \emph{arXiv preprint arXiv:1911.07412}, 2019.

\bibitem[Guo et~al.(2020)Guo, Wang, Li, and Yan]{guo2020dmcp}
Shaopeng Guo, Yujie Wang, Quanquan Li, and Junjie Yan.
\newblock Dmcp: Differentiable markov channel pruning for neural networks.
\newblock In \emph{Proceedings of the IEEE/CVF conference on computer vision
  and pattern recognition}, pages 1539--1547, 2020.

\bibitem[Gao et~al.(2018)Gao, Zhao, Dudziak, Mullins, and Xu]{gao2018dynamic}
Xitong Gao, Yiren Zhao, {\L}ukasz Dudziak, Robert Mullins, and Cheng-zhong Xu.
\newblock Dynamic channel pruning: Feature boosting and suppression.
\newblock \emph{arXiv preprint arXiv:1810.05331}, 2018.

\bibitem[Tang et~al.(2021)Tang, Wang, Xu, Deng, Xu, Tao, and
  Xu]{tang2021manifold}
Yehui Tang, Yunhe Wang, Yixing Xu, Yiping Deng, Chao Xu, Dacheng Tao, and Chang
  Xu.
\newblock Manifold regularized dynamic network pruning.
\newblock In \emph{Proceedings of the IEEE/CVF Conference on Computer Vision
  and Pattern Recognition}, pages 5018--5028, 2021.

\end{thebibliography}


\end{document}